\documentclass[journal]{IEEEtran}
\usepackage{graphics} 
\usepackage{amsmath} 
\usepackage{amssymb}  
\usepackage{amsfonts}
\usepackage{xcolor}
\usepackage{epsfig}
\usepackage{graphicx}
\usepackage{subfigure}
\usepackage{epstopdf} 
\usepackage[linesnumbered,ruled,vlined]{algorithm2e}
\SetNlSty{bfseries}{\color{black}}{}
\usepackage{bm}
\usepackage{gensymb}
\usepackage{verbatim}
\usepackage{cite}
\ifCLASSINFOpdf
\else
\fi

\begin{document}
%
\title{Generative Adversarial Network based Heuristics for Sampling-based Path Planning}
%
%
%

\author{Tianyi Zhang$^{1\dagger}$, Jiankun Wang$^{1\dagger}$ and Max Q.-H. Meng$^{2}$, \emph{Fellow, IEEE}
\thanks{$^{\dagger}$Equal contributions.}
\thanks{$^{1}$Tianyi Zhang and Jiankun Wang are with the Department of Electronic and Electrical Engineering of the Southern University of Science and Technology in Shenzhen, China,
	{\tt\small \{zhangty@mail.,wangjk@\}sustech.edu.cn}}%
\thanks{$^{2}$Max Q.-H. Meng is with the Department of Electronic and Electrical Engineering of the Southern University of Science and Technology in Shenzhen, China, on leave from the Department of Electronic Engineering, The Chinese University of Hong Kong, Hong Kong, and also with the Shenzhen Research Institute of the Chinese University of Hong Kong in Shenzhen, China (\emph{Corresponding authors: }{\tt\small max.meng@ieee.org})}%
}

%


\maketitle

\begin{abstract}
Sampling-based path planning is a popular methodology for robot path planning.
With a uniform sampling strategy to explore the state space, a feasible path can be found without the complex geometric modeling of the configuration space.
However, the quality of initial solution is not guaranteed and the convergence speed to the optimal solution is slow.
In this paper, we present a novel image-based path planning algorithm to overcome these limitations.
Specifically, a generative adversarial network (GAN) is designed to take the environment map (denoted as RGB image) as the input without other preprocessing works.
The output is also an RGB image where the promising region (where a feasible path probably exists) is segmented.
This promising region is utilized as a heuristic to achieve non-uniform sampling for the path planner.
We conduct a number of simulation experiments to validate the effectiveness of the proposed method, and the results demonstrate that our method performs much better in terms of the quality of initial solution and the convergence speed to the optimal solution.
Furthermore, apart from the environments similar to the training set, our method also works well on the environments which are very different from the training set.
\end{abstract}




%
\IEEEpeerreviewmaketitle

\section{Introduction}
\IEEEPARstart{R}{obot} path planning is to determine a collision-free path from a start point to a goal point while optimizing a performance criterion such as distance, time or energy \cite{siciliano2016springer}.
Many classic approaches have been proposed to solve the path planning problem in the past decades, such as potential field method \cite{khatib1986real}, cell decomposition method \cite{lingelbach2004path}, grid-based methods including A*\cite{hart1968formal} and D* \cite{stentz1997optimal} algorithms. 
However, there are some problems existing in these approaches such as computational difficulty and local optimality.

\begin{figure}[t]
	\centering
	\subfigure[RRT*.]{
		\includegraphics[width=40mm]{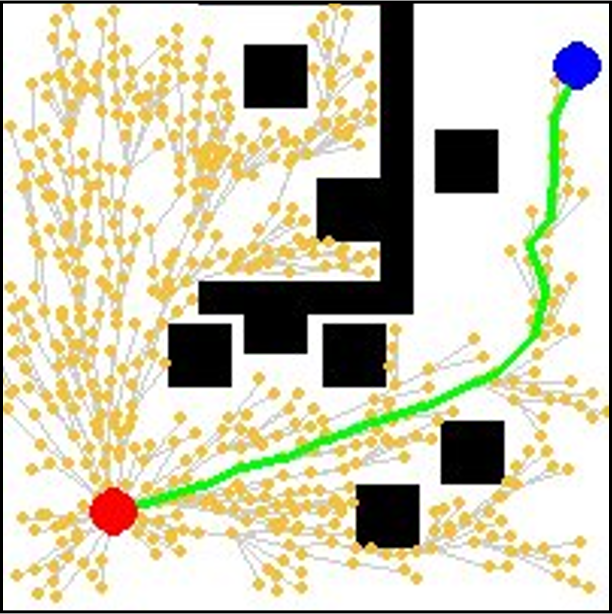}}
	\subfigure[GAN-based heuristic RRT*.]{
		\includegraphics[width=40mm]{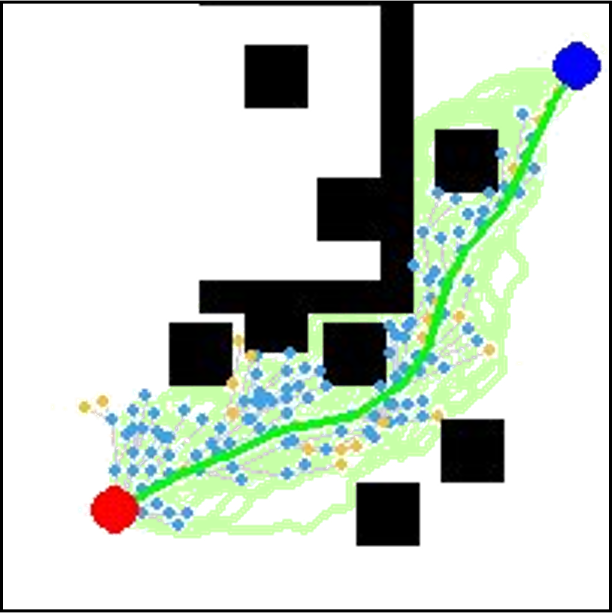}}
	\caption{An comparison of RRT* and GAN-based heuristic RRT* on finding the initial solution. The green area denotes the promising region generated by GAN. The yellow nodes are uniformly sampled from the free state space and the blue nodes are sampled from the promising region, respectively.}
	\label{fig_firstpage}
\end{figure}

In recent years, sampling-based algorithms have been shown to work well in practical applications and satisfy theoretical guarantees such as probabilistic completeness. 
Sampling-based algorithms can avoid discretization and explicit representation of the configuration space, resulting in good scalability to high-dimensional space.
Probabilistic roadmap (PRM) \cite{kavraki1996probabilistic} and rapidly-exploring random tree (RRT) \cite{lavalle1998rapidly} are two representative methods of sampling-based algorithms to solve path planning problems.
RRTs algorithms have gained popularity for their efficient single-query mechanism and easy extension to many applications such as the kinodynamic path planning \cite{webb2013kinodynamic} and socially compliant path planning \cite{wang2019socially}.
However, there are also some limitations of the RRT algorithms which still remain to be solved.
For example, the solutions are usually far from optimal.
Although RRT* \cite{karaman2011sampling} is capable of achieving asymptotic optimality by introducing near neighbor search and tree rewiring operations,  
it takes a lot of time to converge to the optimal solution.
The reason is that the RRT employs a random sampling mechanism.
On one hand, it guarantees that one feasible solution can be found if it exists as the number of iterations goes to infinity.
On the other hand, this mechanism affects the convergence speed because it uniformly searches the whole state space.

To overcome the aforementioned problems, we present a novel learning-based algorithm to leverage previous planning knowledge to generate a promising region where a feasible or optimal path probably exists.
This promising region serves as a non-uniform distribution to heuristically guide the RRT planner to explore the state space more efficiently.
As shown in Fig. \ref{fig_firstpage}, compared with RRT*, our proposed GAN-based heuristic RRT* can quickly find a high-quality initial path with less number of nodes.

Our main contributions include:
\begin{itemize}
	\item Propose a generative adversarial network (GAN) based heuristic method for non-uniform sampling-based path planning;
	\item Design a GAN model to predict the promising region for achieving non-uniform sampling;
	\item Apply the GAN-based heuristic method to RRT* and compare it with the conventional RRT* algorithm, validating the effectiveness of our method;
	\item Test the generative capability of the proposed model, and demonstrate the high success rate and generalization capability.
\end{itemize}

The rest of this paper is organized as follows. 
We first review the previous work of path planning and generative models in Section \ref{related}.
Then, the preliminaries of path planning problem are presented in Section \ref{formulation}.
Section \ref{ModelArchitecture} describes the proposed GAN model for promising region prediction. 
In Section \ref{results}, we conduct a series of simulation experiments to demonstrate the performance of GAN-based heuristic path planing algorithm.
At last, we draw conclusions and talk about the future work in Section \ref{conclusion}.

\section{Related Work}
\label{related}

Recently, many researchers have proposed different types of heuristic methods to improve the performance of basic RRT algorithms. 

As the grid-based algorithms can guarantee to find a \emph{resolution optimal} path, one possible way is to combine the grid-based algorithms with sampling-based ones. 
In \cite{hentoutrrt}, the grid search process is employed to get an initial solution to guide the sampling process.
Another way is to design some specific cost functions to change tree extension rules.
Wang \emph{et al.} \cite{wang2020optimal} utilizes the generalized Voronoi graph (GVG) as the heuristic quality function to quickly extend the tree.
Liu \emph{et al.} \cite{liu2019partition} establish a set of partition heuristic rules for target and collision-free guidance.
While these proposed cost functions accelerate the searching process, they tend to bias the searching towards a certain target, resulting in a local optimum easily or stuck within complex obstacles.
Moreover, the aforementioned methods are task-oriented and only demonstrate fine results on limited conditions, such as low-dimensional spaces or simple cost states.
Their generality and performance cannot be guaranteed in complex environments.
Controlling the sampling distribution is also a useful way to accelerate path planning. 
For example, Gammell \emph{et al.} \cite{gammell2014informed} shrunk the sample space into an elliptic region after a feasible path is obtained.
But its overall performance depends on the quality of initial path.

In recent years, learning-based algorithms show great advantages on path planning problems. 
Especially, they exhibit remarkable generalization to completely unseen environments and can easily be applied to different sizes of maps. 
Baldwin \emph{et al.} \cite{baldwin2010non} propose to learn sampling distributions by using expert data and learn an estimate of sample densities around semantic regions of interest, then incorporate them into a sampling-based planner to produce natural plans.  
Wang \emph{et al.} \cite{wang2018learning} uses a reinforcement learning algorithm to improve the multi-RRT approach in narrow space.
Even though the strategy enhances the local space exploration ability, the expansion of the searching trees is time-consuming.
Ichter \emph{et al.} \cite{ichter2018learning} propose a general methodology for non-uniform sampling based on conditional variational autoencoder (CVAE). 
The weakness of this method is that sampling and planning are separated, hence the planner cannot adapt to the environment during the planning cycle. 
Additionally, the construction of the distributions requires lots of preprocessed conditional information, which demands huge effort on preliminary works. 
Moreover, the method is a multi-process generative model which is time-consuming to predict the whole distribution.
Qureshi \emph{et al.} \cite{qureshi2018deeply} present a neural network-based adaptive sampler to generate samples for optimal paths.
It contains an autoencoder to encode environment from point cloud data and a dropout-based stochastic deep feed-forward neural network to mix the surrounding information and produce samples.
However, their approach only focuses on path generation and require some low-controllers to continuously track the generated paths, which may cause difficulty for robots to act in high-dimensional space.
Wang \emph{et al.} \cite{wang2020neural} proposes a convolutional neural network (CNN) based optimal path planning algorithm to generate a region of interest that is used to guide RRT* sampling process. 
The CNN model is trained on a-prior knowledge and can adapt to different clearances and step sizes. 
However, there are many disconnected parts in the generated regions.

To step further in the learning-based algorithms for controlling the sampling distributions, we can first extract the promising regions of the bias sampling points.
The process of obtaining promising regions can be regarded as a special problem of image semantic segmentation or domain transformation. 
An significant advantage of this image-based generation model is that it does not require any complicated preprocessing.
In the past few years, different semantic segmentation networks are proposed, such as SegNet \cite{badrinarayanan2017segnet}, U-Net \cite{ronneberger2015u}, Mask R-CNN \cite{he2017mask} and so on.
In general, segmentation nets are confined to certain tasks and require labels in the input pictures.
Therefore, the CNN segmentation networks may perform poorly on generating area of interest (ROI). 
Wang \emph{et al.} \cite{wang2020neural} apply segmentation network into path planning problem. 
However, the success rate and connectivity of the promising region is unsatisfactory.

Conditional variational autoencoder (CVAE) \cite{sohn2015learning} is a popular generative model. 
By reparameterization technique, the target images and conditions are encoded and transformed into normal distributions. 
Given the specified condition and random latent variables, the decoder is capable of generating various expected images. 
However, the images generated from CVAE models are often blurred. 
More importantly, it is difficult to restore promising regions from normal distributions. 
Since the advent of GANs \cite{goodfellow2014generative}, it has been acknowledged as the most powerful generative model and widely used in various areas, such as image generation \cite{lu2018image} \cite{han2018gan} \cite{zhai2019lifelong}, image-to-image translations \cite{cherian2019sem} \cite{li2019asymmetric} \cite{guo2020gan} and so on. 
Due to the flexibility of GANs, different architectures are proposed in recent years to tackle diverse problems. 
It is also suitable for generating promising regions of non-uniform sampling.
Therefore, we design a novel GAN structure to learn promising regions for achieving heuristic non-uniform sampling to improve the performance of RRT*.
Compared with the previous methods, our image-based model avoids complicated preprocessing works of the state space and shows good adaptability to various environments. 
   
\section{Preliminaries}
\label{formulation}
In this section, we first briefly introduce the path planning problem and then formulate our proposed heuristic based RRT* algorithm.

\subsection{Path Planning Problem}
\label{PathPlanningProblem}
The basic path planning problem can be defined as follows.
Let $\mathcal{X} \in \mathbb{R}^n$ be the state space, with $n \in \mathbb{N}^n$, ${n} \geq 2$.
The obstacle space and the free space are denoted as $\mathcal{X}_{obs}$ and $\mathcal{X}_{free} = \mathcal{X} / \mathcal{X}_{obs}$, respectively.
Let $x_{init}$ be the initial state and $x_{goal}$ be the goal state where $x_{init} \in \mathcal{X}_{free} $, $ x_{goal} \in \mathcal{X}_{free}$. 
Let $\mathcal{X}_{goal}$ be the goal region where $\mathcal{X}_{goal} =\{x \in \mathcal{X}_{obs} \big| || x - x_{goal} || < r\} $.
The path planning aims to find a feasible path $\sigma: [0, 1] \rightarrow \mathcal{X}_{free}$,  $\sigma(0) = x_{init}$ and $\sigma(1) \in \mathcal{X}_{goal}$.

Let $\Sigma$ be the set of feasible paths and $c(\sigma)$ be the cost function.
The optimal path planning problem is to find a path $\sigma^*$ with the lowest cost in $\Sigma$:
\begin{align}
	\label{Eq_pathplanning}
	\sigma^* = \ & \underset{\sigma \in \Sigma}{\arg\min} \ c(\sigma) \nonumber \\
	s.t. \     & \sigma(0) = x_{init} \nonumber \\
	& \sigma(1) \in \mathcal{X}_{goal} \nonumber \\
	& \sigma(t) \in \mathcal{X}_{free}(t), \forall t \in [0,1].
\end{align}
In this paper, the cost function between two state $x_{i}$ and $x_{j}$ is defined as
\begin{align}
	\label{Eq_PathPlanningCost}
	Cost(x_i,x_j) = || x_{i} - x_{j} ||.
\end{align}

\subsection{Heuristic based RRT*}
\label{ProblemFormulation}
Sampling-based algorithms solve path planning problems through constructing space-filling trees to search a path $\sigma$ connecting start and goal state. 
The tree is built incrementally with samples drawn randomly from the free space $\mathcal{X}_{free}$.

To reduce unnecessary sampling, we construct a heuristic non-uniform sampling distribution $\mathcal{X}_{H} \subset \mathcal{X}_{free} $ to improve the sampling process. 
A heuristic non-uniform sampling distribution refers to the state space where feasible paths exist with high probability. 
In order to obtain this non-uniform sampling distribution, we first generate a promising region with GAN model, and then discretize it.
The outline of our heuristic methodology is shown in Alg. \ref{Algs.outline}.

\begin{algorithm}[h]
	\DontPrintSemicolon
	\SetKwRepeat{Do}{do}{while}
	\SetKwInOut{Input}{Input}
	\SetKwInOut{Output}{Output}
	\Input{$x_{init}$, $x_{goal}$, $Map$}
	\Output{$G(V,E)$}
	$V = x_{init},  E = \emptyset;$\;
	$\mathcal{S} \leftarrow  \text{ROIGenerator}(x_{init}, x_{goal}, Map);$\;
	$\mathcal{X}_{H}  \leftarrow \text{Discretization}(\mathcal{S});$\;
	$G(V,E) \leftarrow \text{HeuristicSBP} \text{*} (x_{init}, x_{goal}, Map,\mathcal{H});$\;
	$\text{Return} \ G(V,E);$\;
	\caption{Outline of GAN-based heuristic RRT* \label{Algs.outline}}
\end{algorithm}

In this paper, we formulate our model in complex $2$-dimensional environments. 
As for the sampling-based algorithm, we employ the most general optimal path planning method RRT* \cite{karaman2011sampling} as a representative.
Alg. \ref{Algs.RRT*} illustrates the detailed implementation of our heuristic RRT* algorithm and compares it with basic RRT*.
Line $3$ to Line $7$ is adopted by heuristic RRT* and Line $8$ to Line $9$ is in RRT*.
We modify the uniform sampling strategy into a biased sampling strategy: in heuristic RRT*, $\mu\%$ samples are randomly chosen from the non-uniform sampling distribution, while $1-\mu\%$ are sampled from the uniform sampling distribution to guarantee probabilistic completeness.
Here we denote $\mathcal{X}_H$ as heuristic discrete points extracted from the sampling distribution.

\begin{algorithm}[h]
	\DontPrintSemicolon
	\SetKwRepeat{Do}{do}{while}
	\SetKwInOut{Input}{Input}
	\SetKwInOut{Output}{Output}
	\Input{$x_{init}$, $\mathcal{X}_{goal}$,$\mathcal{H}$, $Map$ and $Use Heuristic$}
	\Output{$G(V,E)$}
	$V = x_{init},  E = \emptyset;$\;
	\For{$i = 1 \cdots N$}{
		\color{black}
		\If {$Use Heuristic = True$}{
		\If {$\text{Rand}() > \mu$}{
			$x_{rand} \leftarrow \text{Non-uniformSample}(\mathcal{X}_H);$\;	
		}
		\Else{
			$x_{rand} \leftarrow\text{UniformSample}();$\;
		}	
	}
		\color{black}
		\Else{
			$x_{rand} \leftarrow \text{UniformSample()};$\;
		}
		
		\color{black}
		$x_{nearest} \leftarrow \text{Nearest}(G,x_{rand});$\;
		$x_{new} \leftarrow \text{Steer}(x_{nearest},x_{rand});$\;
		
		\If {$\text{ObstacleFree}(x_{nearest},x_{rand})$}{
			$\text{Extend}(G,x_{new});$\;
			$\text{Rewire}();$\;
			\If{$x_{new} \in \mathcal{X}_{goal}$}{
				$\text{Return} \ G(V,E);$\;}}
		$\text{Return} \ failure;$\;
	}
	\caption{Comparison of RRT* and Heuristic RRT* \label{Algs.RRT*}}
\end{algorithm}

To sum up, the focus of our work lies in establishing an efficient generator to predict promising region $\mathcal{S}$ under the given conditions $x_{init}, x_{goal}, Map$ (shown in Alg. \ref{Algs.outline} Line $2$).

\section{GAN-based Promising Region Generation}
\label{ModelStructure}
In this section, we introduce the proposed GAN model in details.
Trained with large amount of empirical promising region data, the GAN model is able to generate promising regions for non-uniform sampling under specified conditions.
The input of the model is an RGB image representing the state space $Map$, start state $x_{init}$, and goal state $x_{goal}$.  
The output of the model is also an RGB image where the promising regions are highlighted. 

\subsection{Dataset Generation}
\label{DataGeneration}

\begin{figure}[h]
	\centering
	\includegraphics[width=88mm]{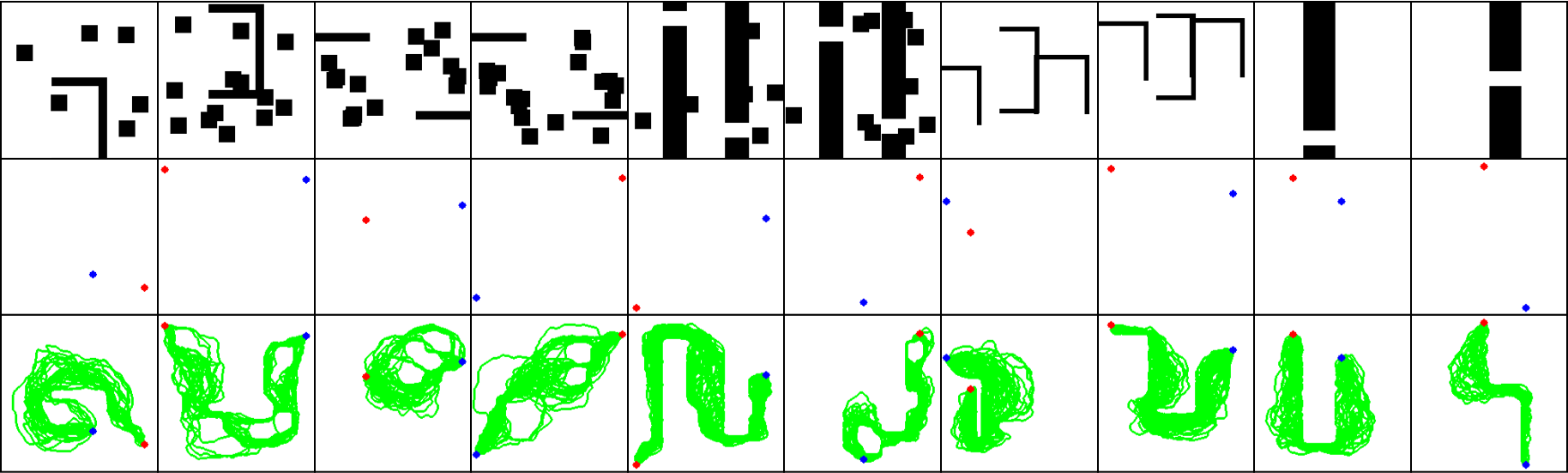}
	\caption{An illustration of the dataset. Each row from top to bottom represents maps, points, and promising regions.}
	\label{fig_dataset}
\end{figure}

Fig. \ref{fig_dataset} illustrates dataset.
Instead of coordinates, we use an image of the same size as the map to represent the coordinate information.
The environment maps ($201 \times 201$ pixels) are randomly generated. 
In the environment map, the obstacles are denoted as black and the free spaces are denoted as white, respectively.
On each map, $20$ to $50$ start and goal states are randomly chosen from the free space.
They are denoted as red and blue, respectively.
To obtain the ground truth for training,  we use the set of feasible paths to represent the promising region on each specific condition, as shown in Fig. \ref{fig_dataset}. 
Specifically, for each condition, the start and goal state are randomly generated, and the environment map is also randomly generated.
We run $50$ times RRT algorithm on each condition and draw all the feasible paths on the image with green lines.

\subsection{Model Structure}
\label{ModelArchitecture}

\subsubsection{Theory of GAN}
\

We first briefly introduce the theory of GAN and then discuss the connection between our model and the theory. 
The basic framework of GAN includes a generator $G$ and a discriminator $D$. 
The generator accepts a sample noise $z$ from the noise space $\mathcal{Z} \in \mathbb{R}^n $ and outputs an image $G(z)$. 
The discriminator is fed with images $u$ from the training data space $\mathcal{U}$ and generates images $G(z)$.
The generator and discriminator are in an adversarial game: to be specific, the generator aims at fooling the discriminator while the discriminator tries to distinguish the real and fake images.
The objective function can be defined as
\begin{equation}
\begin{split}
	\underset{G} {\min} \ \underset{D}{\max} \ &
	(\mathbb{E}_{u \sim P_{data}(U)}[\log D(u)] + \\ 
	& \mathbb{E}_{z \sim P_{\mathcal{Z}}(z)}[\log (1-D(G(z)))]).
	\label{eq_gan}
\end{split}
\end{equation}
To control the generative process, the generator and discriminator can be conditioned on a given $y \in \mathcal{Y}$, where $\mathcal{Y}$ is the condition space.
In this way, the objective function is modified into 
\begin{equation}
\begin{split}
	\underset{G} {\min} \ \underset{D}{\max} \ &
	(\mathbb{E}_{u,y \sim P_{data}(u,y)}[\log D(u,y)] + \\
	& \mathbb{E}_{z \sim P_{\mathcal{Z}}(z),y \sim P_{\mathcal{y}}(y)}[\log (1 - D(G(z,y),y))]).
	\label{eq_cgan}
\end{split}
\end{equation}

During the training process, we first label $\mathcal{U}$ as real and $G(\mathcal{Z})$ as fake.
Then the discriminator and generator are trained alternatively, which leads to the mutual growth of discernibility and generative capability. 

\begin{figure}[t]
	\centering
	\includegraphics[width=88mm]{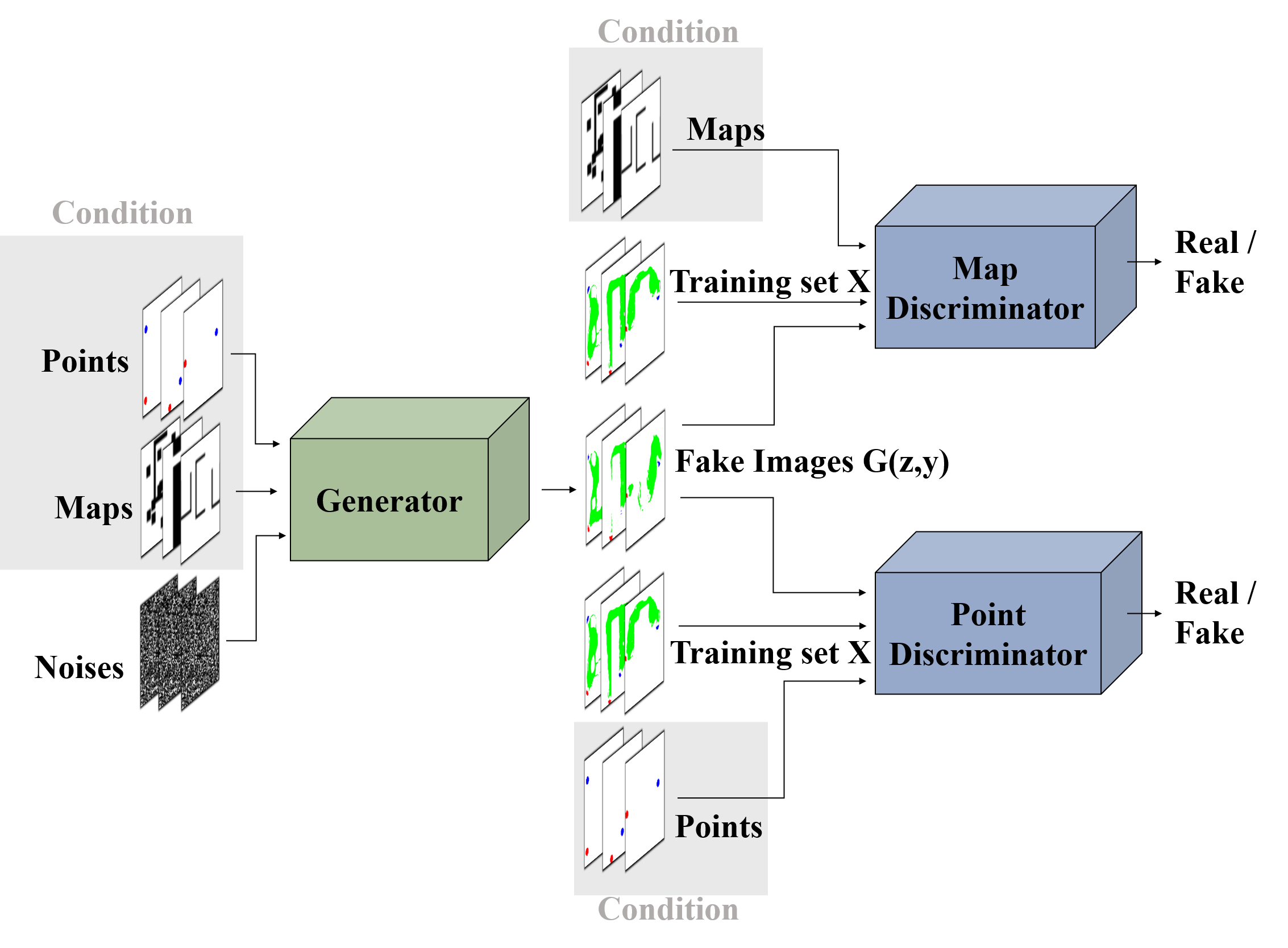}
	\caption{The overall structure of GAN model for promising region generation.}
	\label{fig_ModelStructure1}
\end{figure}

In our model, we refer to promising region images $\mathcal{S}$ as data space $\mathcal{U}$.
Map images $\mathcal{M}$ and points images (start and goal state) $\mathcal{P}$ are two conditions in space $\mathcal{Y}_{map}$, $\mathcal{Y}_{point}$ and $\mathcal{Z}$ is $2$-D noise data. 
The task is to parameterize $G$ to fit the distribution of $G(\mathcal{Z})$ to $\mathcal{U}$.
To enhance the network's ability to locate start and goal points, we use two discriminators $D_{map}$ and $D_{point}$ to judge whether the promising regions match the map and point conditions, respectively.
Fig. \ref{fig_ModelStructure1} shows the overall architecture of our model. 

\begin{figure*}[t]
	\centering
	\includegraphics[width=180mm]{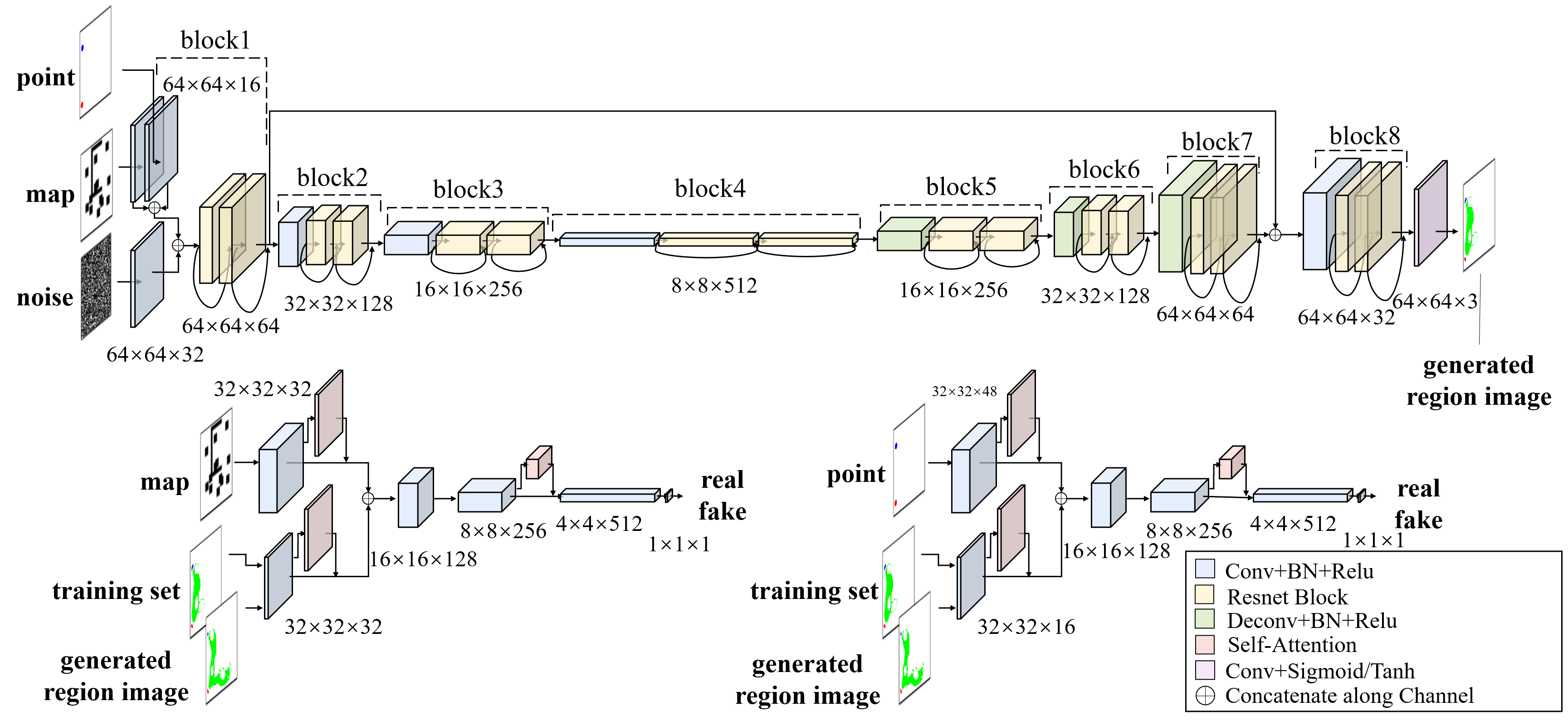}
	\caption{The detailed structure of GAN model for promising region generation.}
	\label{fig_ModelStructure2}
\end{figure*}

\subsubsection{Loss Function}
\ 

Eq. \ref{eq_gan} and Eq. \ref{eq_cgan} show the object function of generative adversarial networks. 
Herein, we introduce their implementations in the loss functions.

During the training process, we first fix the parameters in the generator and train the discriminators for $2$ iterations. 
The loss function of discriminators can be defined as
\begin{align}
	\label{Eq_LossFunction_Discriminator}
	& \mathcal{L}_{D_{map}}  =  \ 
	\mathbb{E}[\log D_{map}(s,m)] +  \nonumber \\ 
	& \qquad \qquad \ \ \mathbb{E}_[\log (1 - D_{map}(G(z,m,p),m))] \nonumber \\
	& \mathcal{L}_{D_{point}}  =  \ 
	\mathbb{E}[\log D_{point}(s,p)] +  \nonumber \\ 
	& \qquad \qquad \ \ \mathbb{E}_[\log (1 - D_{point}(G(z,m,p),p))] \nonumber \\
	& \qquad s.t.\ s,m,p \sim P_{\mathcal{S,M,P}}(s,m,p), z \sim P_{\mathcal{Z}}(z).
\end{align}
Then we train the generator with the following loss function:
\begin{equation}
	\begin{split}
		\mathcal{L}_{G}  = \ & \alpha_1 \mathbb{E}[\log D_{map}(G(z,m,p),m)] + \\ 
		\ & \alpha_2 \mathbb{E}[\log D_{point}(G(z,m,p),p)] \\
		& s.t.\ m,p \sim P_{\mathcal{M,P}}(m,p), z \sim P_{\mathcal{Z}}(z).
		\label{Eq_LossFunction_Generator1}
	\end{split}
\end{equation}
As the start and goal state occupy small pixels in the images, the generator may ignore the semantic information of them. 
To enhance generator's attention to start and goal states, We design dynamic cross coefficients $\alpha_1$ and $\alpha_2$ to give larger weight to the loss of $D_{point}$.
The specific calculation formulas are represented in Eq. \ref{Eq_LossFunction_Generator2}.
$k$ is a hyperparameter. 
In our model, we set $k$ to $3$.
    
\begin{equation}
	\begin{split}
		\mathcal{L}_{G}  = &  \ 
		\frac{k \mathcal{L}_{D_{map}}}{\mathcal{L}_{D_{point}} + k \mathcal{L}_{D_{map}}} \mathbb{E}[\log D_{map}(G(z,m,p),m)] + \\ 
		& \frac{ \mathcal{L}_{D_{point}}}{\mathcal{L}_{D_{point}} + k \mathcal{L}_{D_{map}}} \mathbb{E}[\log D_{point}(G(z,m,p),p)] \\
		& s.t.\ m,p \sim P_{\mathcal{M,P}}(m,p), z \sim P_{\mathcal{Z}}(z).
		\label{Eq_LossFunction_Generator2}
	\end{split}
\end{equation}
 
\subsubsection{Detailed Model Structure}
\

Fig. \ref{fig_ModelStructure1} shows the overall architecture the GAN model for promising region generation and Fig. \ref{fig_ModelStructure2} illustrates the detailed structure of the generator and discriminators.

In the generator, the inputs are $64\times 64 \times 1$ dimensional noise image $\mathcal{Z}$, $64 \times 64 \times 3$ dimensional map image $\mathcal{M}$ and $64 \times 64 \times 3$ dimensional points image $\mathcal{P}$.
The output is the $64 \times 64 \times 3$ dimensional promising region image $\mathcal{S}$.
$\mathcal{M}$ and $\mathcal{P}$ are separately fed into an $16$-channel convolutional layer and $\mathcal{Z}$ is fed into an $32$-channel convolutional layer, respectively.
Then the feature maps are concatenated together along channel dimension and processed by the subsequent networks to encode the images into multi-channel features.

In our model, the generator obeys the encoder-decoder architecture.
The encoding network contains $block1$ to $block4$ and each block is composed by an Convolution-BatchNormalization-ReLU layer and two resnet cells \cite{he2016deep}. 
Herein, we store the feature maps derived from $block1$ and denote it as $I_1$. 
The decoding network includes $block5$ to $block7$. 
Each decoder block is similar to the encoder ones, only replacing the convolutional layers with deconvolutional layers. 
Likewise, we denote the output feature maps of decoding network as $I_2$.
$Block8$ is designed to improve the image quality by enhancing raw context information. 
The feature maps $I_1$ and $I_2$ are concatenated along channel and fed into $block8$.
At the end of the network, the feature maps are compressed into $3$-channel image and activated by Tanh function.

The two discriminators have similar structure except for the number of the first channels. 
The discriminators are mainly constructed by Convolution-BatchNormalization-LeakyReLU blocks.
Inspired by self-attention mechanism\cite{wang2018non}\cite{zhang2019self}, we add self-attention blocks at the first layer and third layer of the discriminators to help them consider global information.

As shown in Fig. \ref{fig_ModelStructure2}, self-attention blocks are added to the convolutional network through residual connection:
\begin{equation}
	\begin{split}
		\mathbf{o}_a = \gamma \mathbf{q} + \mathbf{o}_c
	\end{split}
\end{equation}
where $\mathbf{q}$ refers to the output of self-attention mechanism, $\mathbf{o}_c$ denotes the output of the last layer, and $\mathbf{o}_a$ is the final output.
$\gamma$ is a learned parameter to adjust the ratio of self attention. 
The self attention mechanism can be defined as 
\begin{equation}
	\begin{split}
		\mathbf{q} & = softmax(\theta(\mathbf{o}_c)^T  \phi(\mathbf{o}_c)) g(\mathbf{o}_c) \\ &
		 = softmax(\mathbf{o}_c^T W_{\theta}^{T}  W_{\phi} \mathbf{o}_c) W_{g}\mathbf{o}_c
	\end{split}
\end{equation}
where $g(\mathbf{o}_c)$ is a linear embedding used to compress dimensions. 
$\theta(\mathbf{o}_c)$ and $\phi(\mathbf{o}_c)$ are used to calculate autocorrelation.
Through $softmax$ operation, the weight coefficients are derived and added to $g(\mathbf{o}_c)$. 
In this way, all positions are considered in the operation.

In the discriminators, the condition input channel is fed with $64 \times 64 \times 3$ dimensional maps $\mathcal{M}$ or point images $\mathcal{P}$ and the data input channel receives promising region images $\mathcal{S}$.
When data $\mathcal{S}$ is obtained from the training set, we label the output of discriminators as $1$ to mark it as real. 
When data $\mathcal{S}$ is generated from the generator, we label the output as $0$ to mark it as fake.

\section{Experiment Results}
\label{results}

\subsection{Training Details}
\label{TrainingDetails}
We generate $61768$ sets of data and randomly choose $49416$ among them for training and $3088$ for test. 
As shown in Fig. \ref{fig_dataset}, we mainly use five types of map sets, each of which contains $100$ to $300$ different environment maps. 
On each environment map, we randomly choose $20$ to $50$ different start and goal states. 
The training process is conducted for $50$ epochs on NVIDIA TESLA T4 with PyTorch. 
We use Adam optimizer with parameters $\beta_1=0.5$, $\beta_2=0.99$ for training, and the learning rate of discriminators and generator are $0.00001$ and $0.0001$, respectively. 

\subsection{Evaluation Methods}
\label{EvaluationMethods}

We evaluate our model from two aspects.
As an image generation problem, we evaluate the connectivity of the generated promising regions and generalization ability of the model. 
As a path planning problem, we apply the promising region into Alg. \ref{Algs.RRT*} to compare several standard metrics between basic RRT* and our GAN-based heuristic RRT*.

\subsubsection{Connectivity}
\ 

We use connectivity to evaluate the image quality instead of other image similarity measure metrics, such as histogram-based method, SSIM, L1 loss, etc.
Specifically, we set the promising region (green area) as free space $X_{free(test)}$ and denote it as white.
Then we execute RRT algorithm in $X_{free(test)}$: if feasible paths can be found, it means that $X_{free(test)}$ is connected.

\begin{figure}[h]
	\centering
	\subfigure[Unsuccessfully connected.]{
		\includegraphics[width=87mm]{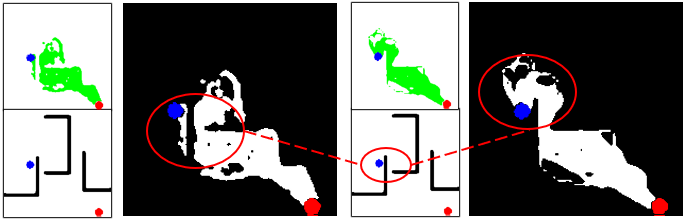}}
	\subfigure[Successfully connected.]{
		\includegraphics[width=88mm]{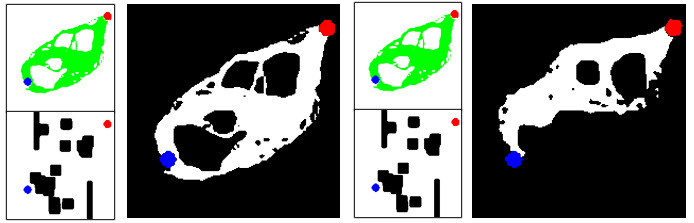}}
	\caption{An illustration of the method for connectivity evaluation. In each subfigure, the left set of images are from the training set and the right images are the generated ones.}
	\label{fig_connectivity}
\end{figure}

Fig. \ref{fig_connectivity} displays the process of our task-based connectivity evaluation method and the advantages of it.
In Fig. \ref{fig_connectivity}(a), although the shape of the generated image and the image in the training set are very similar, the quality of the generated image is not good.
In Fig. \ref{fig_connectivity}(b), even though the generated image looks very different from the image in the training set, the quality of the generated image is good because the start and goal points are well connected.

\subsubsection{Generalization Ability}
\ 

\begin{figure}[t]
	\centering
	\includegraphics[width=88mm]{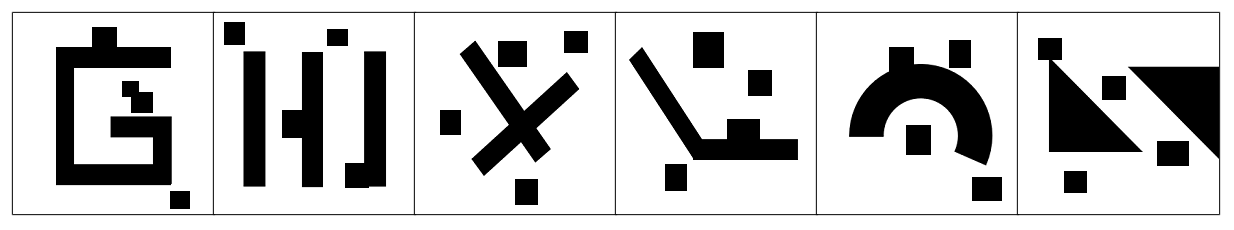}
	\caption{Examples of completely new environment maps for generalization ability test.}
	\label{fig_unseen_map}
\end{figure}

Another measurement for model quality is the generalization ability in completely different environments.  
As shown in Fig. \ref{fig_unseen_map}, we generate $6$ types of different map sets which have not been seen before.
As the method shown in \ref{DataGeneration}, we randomly set 20 to 50 pairs of start and goal points and run RRT to generate ground truth promising regions for comparison.
The new dataset contains $1200$ images.

\subsubsection{Improvement for RRT*}
\ 

To test the improvement of the proposed GAN-based promising region generation on sampling-based planning algorithms, we extract the heuristic non-uniform sampling distribution from the promising region and proposed the heuristic based RRT* (shown in Alg. \ref{Algs.outline} and Alg. \ref{Algs.RRT*}).
To highlight the influence of GAN-based heuristic and guarantee the probabilistic completeness simultaneously, we choose the hyperparameter $\mu$ in Alg. \ref{Algs.RRT*} as $10$, which means that $90\%$ nodes are randomly chosen from the heuristic non-uniform sampling distribution and $10\% $ are chosen uniformly from the free space. 

We compare the improvement on the process of finding a feasible path and converging to a optimal path through three metrics, including path cost, number of nodes, and planning time.
Path cost represents the quality of the path, and we use Euclidean distance as the cost function (shown in Eq. \ref{Eq_PathPlanningCost}). 
The number of nodes indicates the requirement of memory space and planning time refers to the time consumption of finding a feasible path or an optimal path.
These two indexes reflect the availability of the algorithm when applied into practical applications. 

\begin{figure}[h]
	\centering
	\includegraphics[width=100mm]{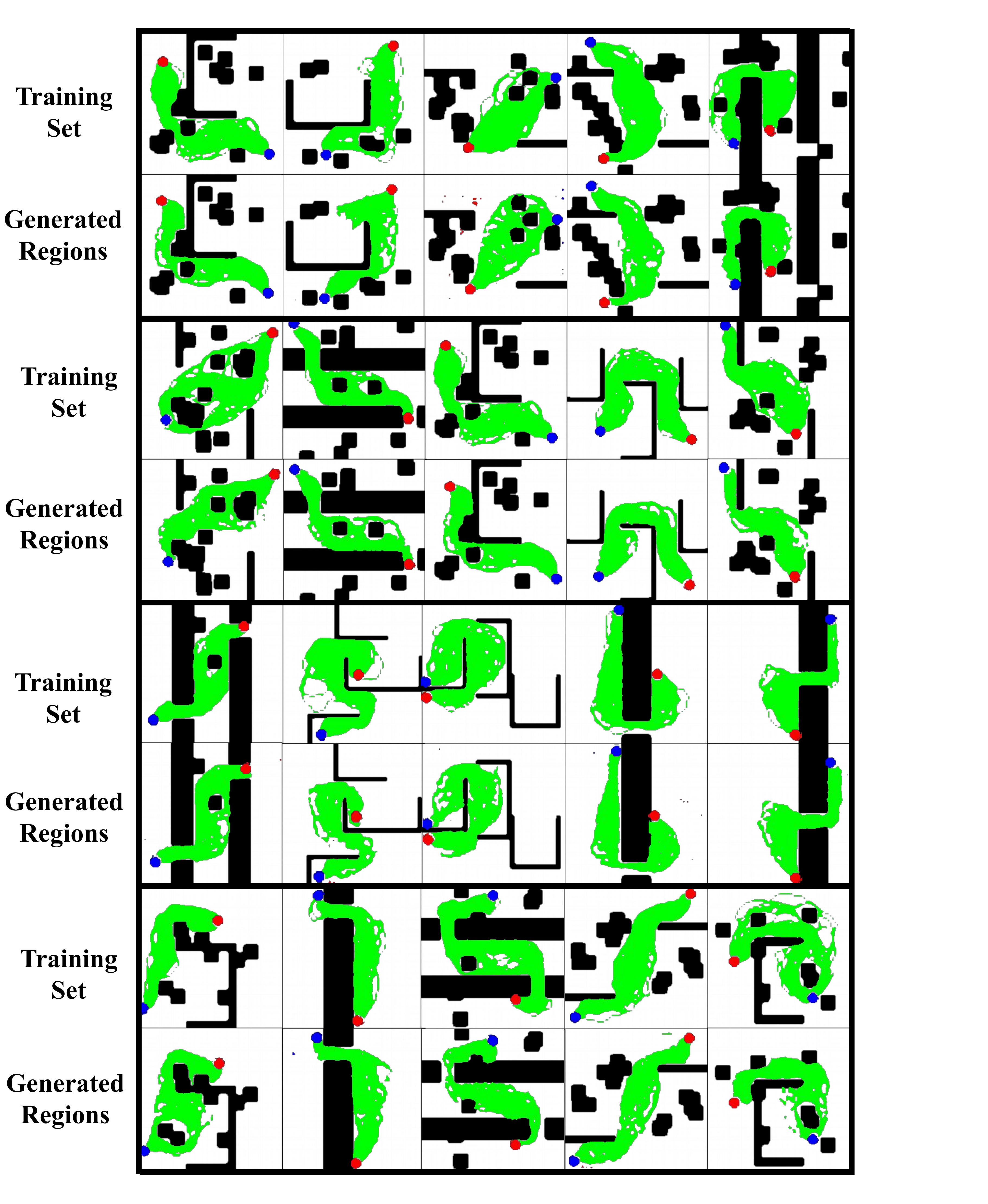}
	\caption{Examples of generated promising region images in the test set. The upper images are from the training set and the lower images are from the generator.}
	\label{fig_testset_result}
\end{figure}
 
\subsection{Experiment Results}
\label{ExperimentResult}

\subsubsection{Connectivity}
\ 

In our experiment, the success rate reaches $89.83\%$.
Some of the results are represented in Fig. \ref{fig_testset_result}.
It demonstrates that our model can generate high-quality continuous promising regions on different conditions.

\subsubsection{Generalization Ability}
\ 

\begin{figure}[h]
	\centering
	\includegraphics[width=85mm]{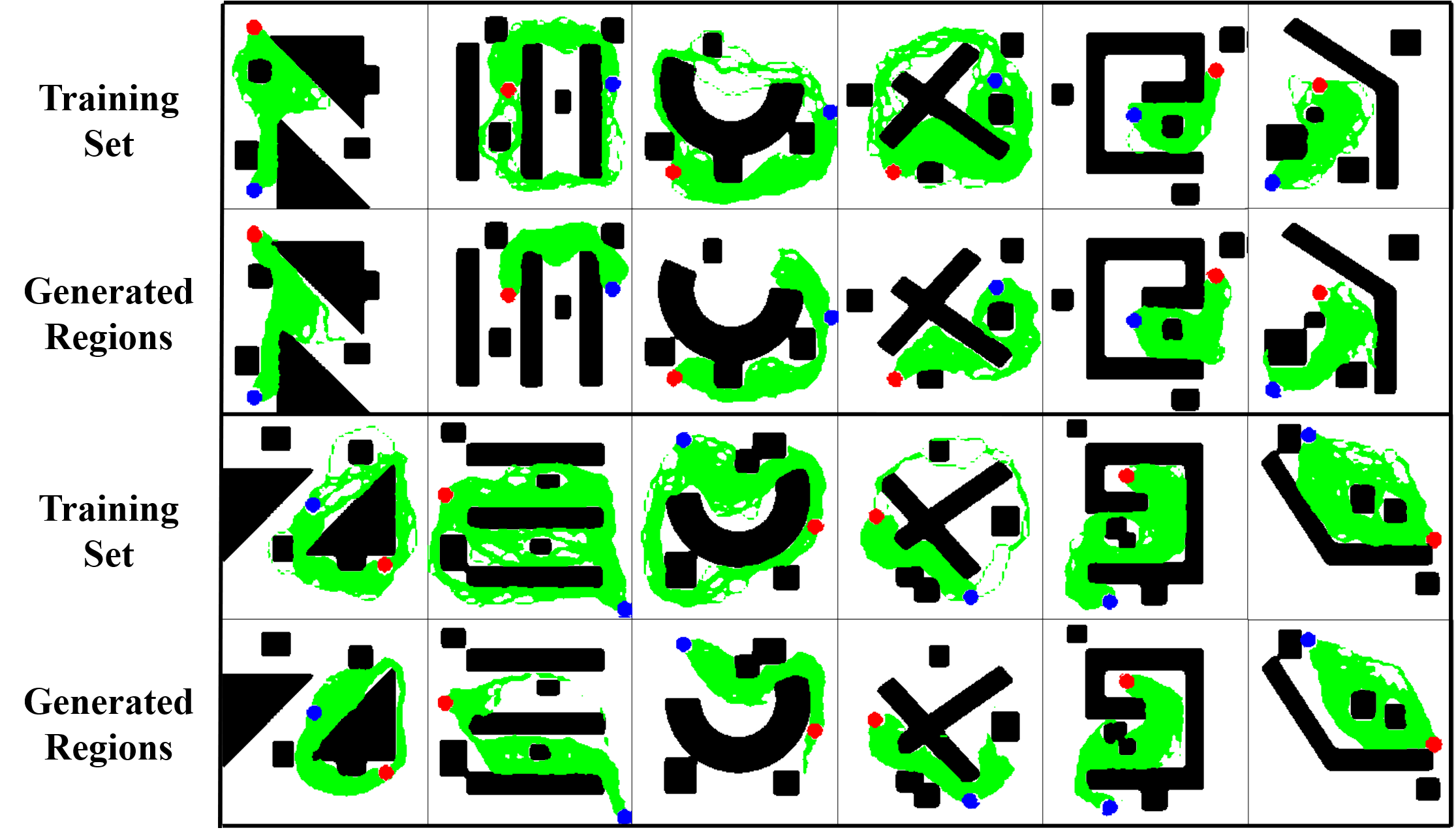}
	\caption{Examples of generated promising region image in the unseen test set. The upper images are from the training set and the lower images are from the generator.}
	\label{fig_unseen_result}
\end{figure}

As shown in Fig. \ref{fig_unseen_result}, our model shows good adaptability on unseen environments.
It predicts accurate continuous promising regions on totally unseen maps and the success rate can achieve $81.9\%$. 

\subsubsection{Improvement on RRT*}
\ 

\begin{figure}[h]
	\centering
	\includegraphics[width=85mm]{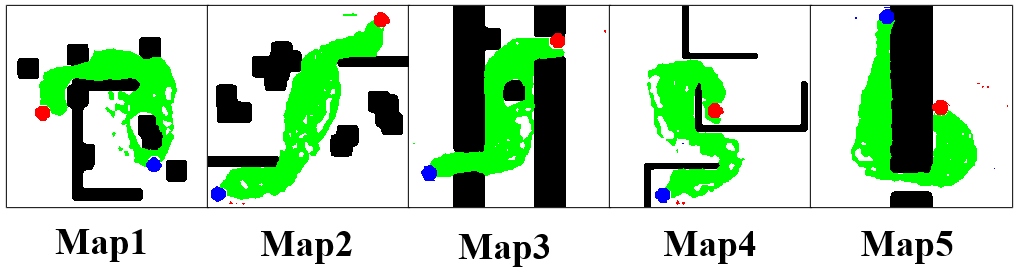}
	\caption{Five conditions for testing the improvement of GAN-based heuristic RRT*.}
	\label{fig_testmap}
\end{figure}

To illustrate the improvement on RRT*, we randomly choose one result from each type of the maps in the test set, which is represented in Fig. \ref{fig_testmap}.
As RRT* have strong randomness, we run GAN-based heuristic RRT* and basic RRT* for $50$ times on Python 3.8 to get the statistics of evaluation results. 

\textit{a. Initial Stage}

\begin{figure}[h]
	\centering
	\subfigure[Initial searching on Map1.]{
		\includegraphics[width=85mm]{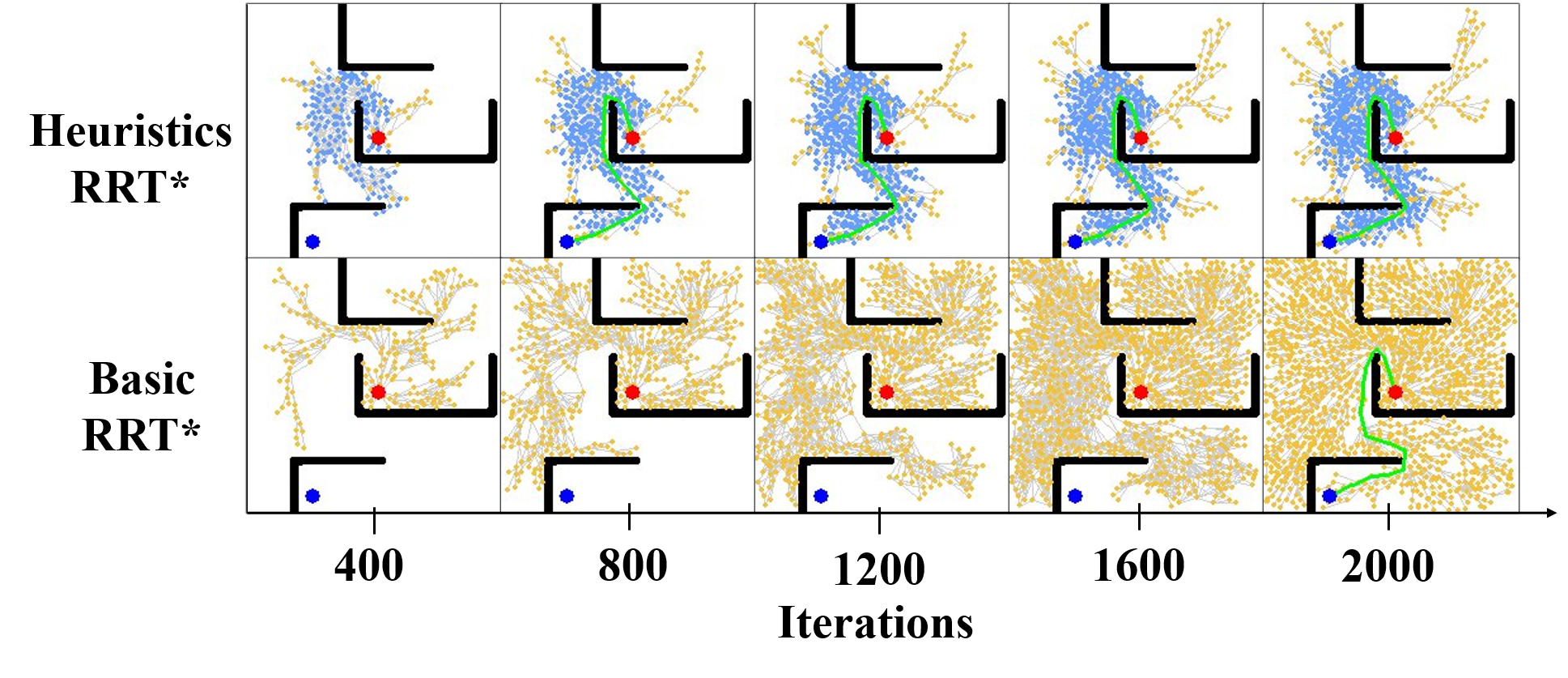}}
	\subfigure[Initial searching on Map4.]{
		\includegraphics[width=85mm]{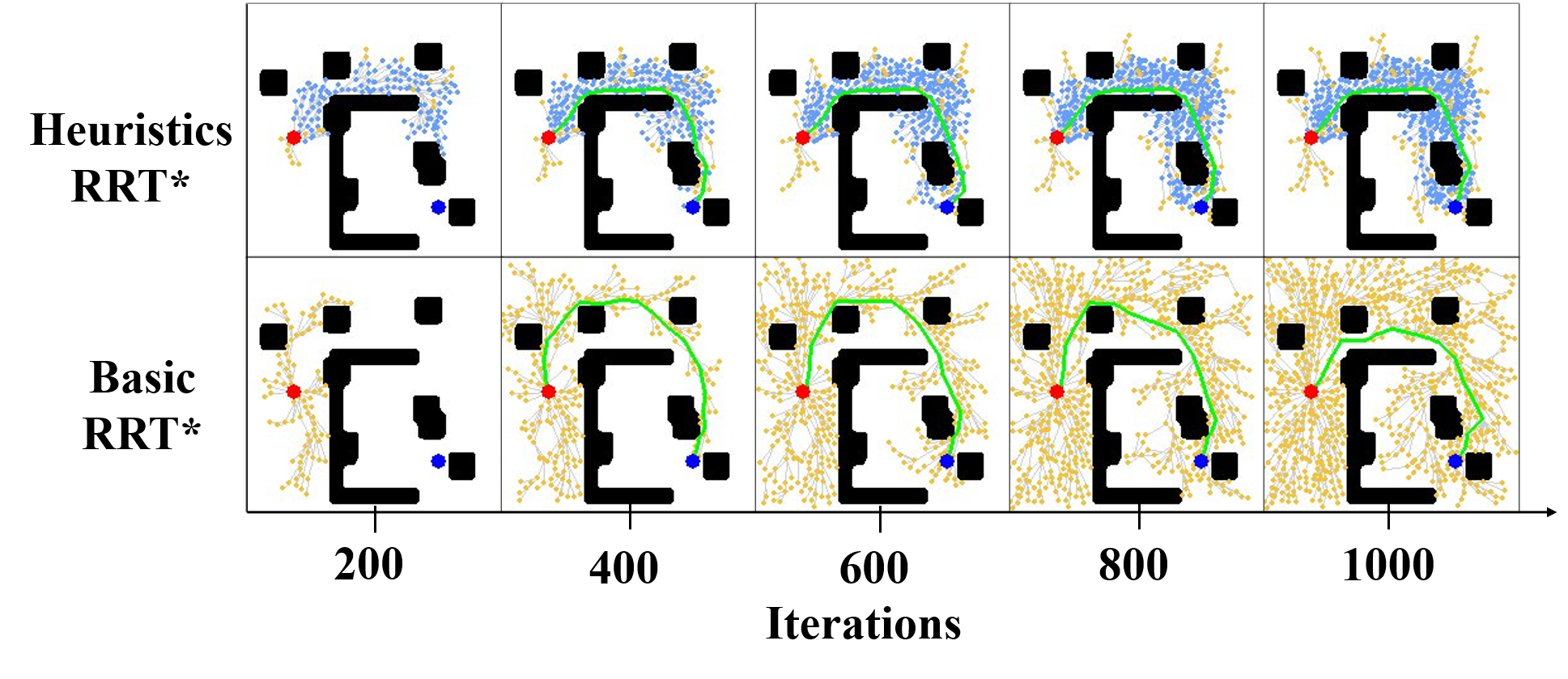}}
	\caption{Comparison of the searching process between GAN-based heuristic RRT* and basic RRT* at the initial stage of planning. The X-axis refers to the number of searching iteration in RRT*. The yellow nodes are uniformly sampled from the free state space and the blue nodes are sampled from the promising region, respectively. The initial path is marked as green.}
	\label{fig_initial_expansion}
\end{figure}

The initial stage refers to the period from the beginning to the moment when the first feasible path is found.
Fig. \ref{fig_initial_expansion} displays the initial expansion of RRT* trees. 
In Fig. \ref{fig_initial_expansion}(a), the non-uniform sampling distribution guides RRT* to search in the space with high possibility of optimal paths, resulting in lower path cost within the initial planning stage.
Fig. \ref{fig_initial_expansion}(b) represents a more difficult task and highlights the time advantage of heuristic RRT*. 
The result shows that the RRT* needs more $1352$ iterations than GAN-based heuristic RRT* to find the goal.

\begin{figure*}[t]
	\centering
	\subfigure[Initial path cost.]{
		\includegraphics[width=58mm]{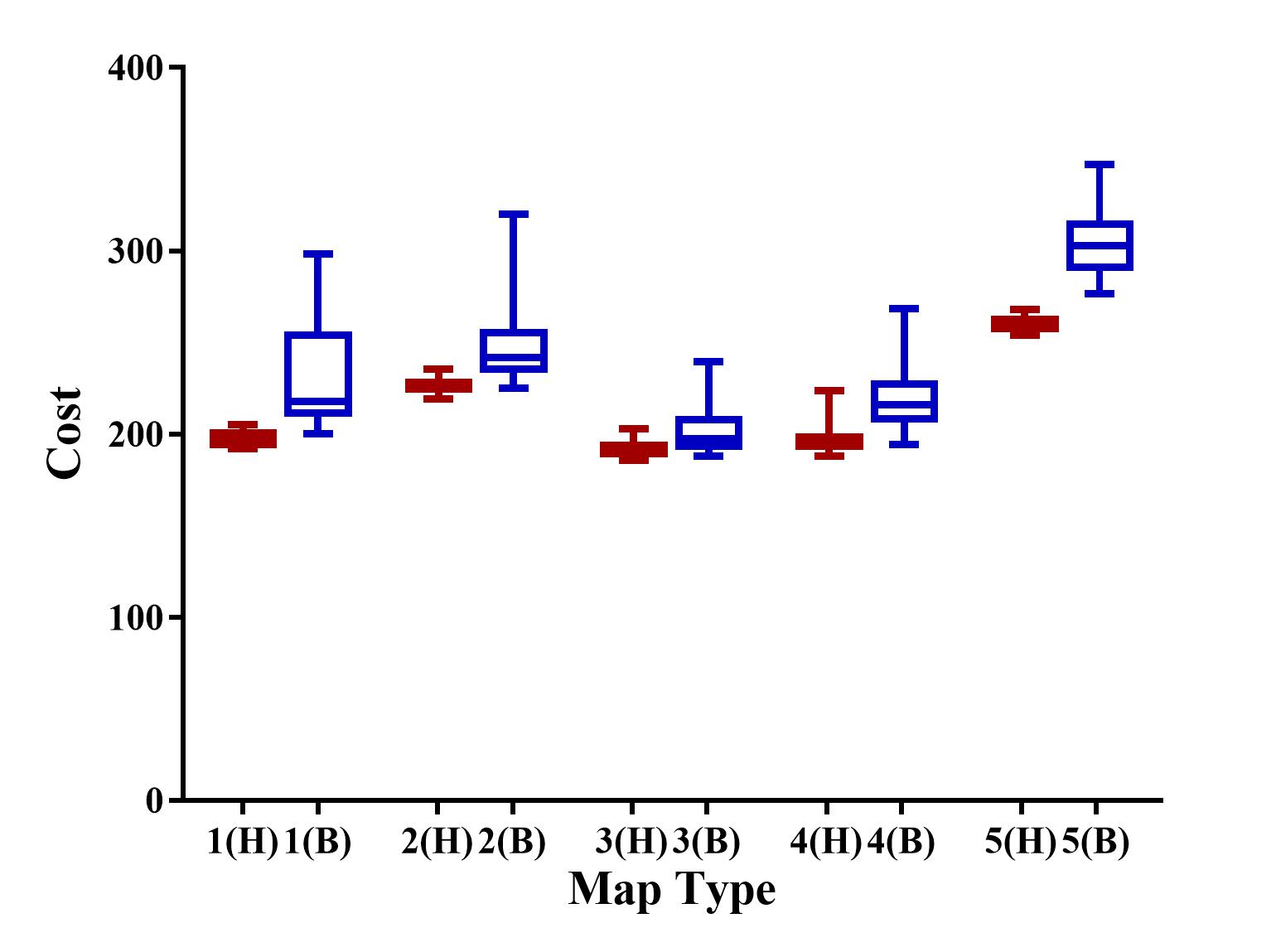}}
	\subfigure[Initial time cost.]{
		\includegraphics[width=58mm]{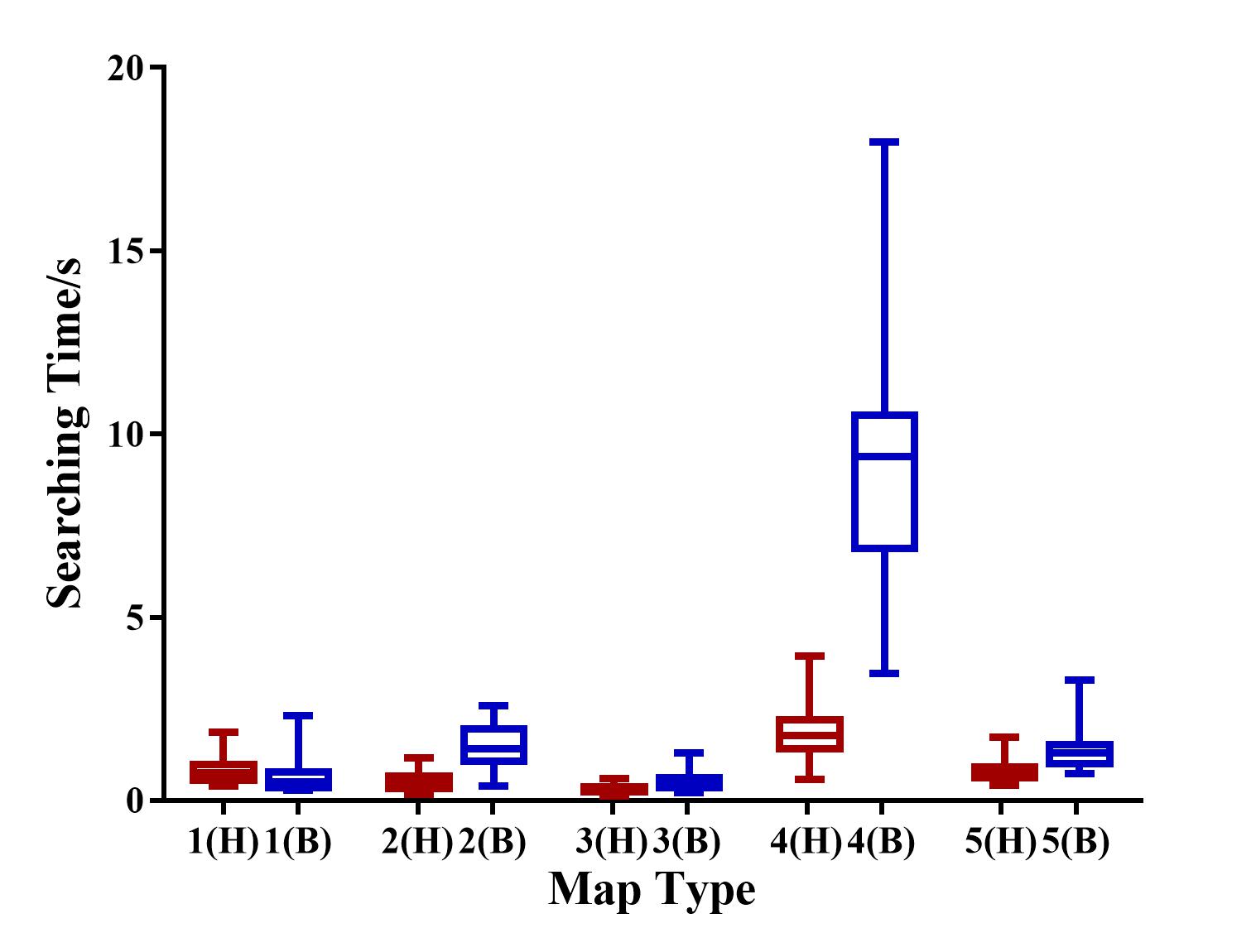}}
	\subfigure[Initial nodes.]{
		\includegraphics[width=58mm]{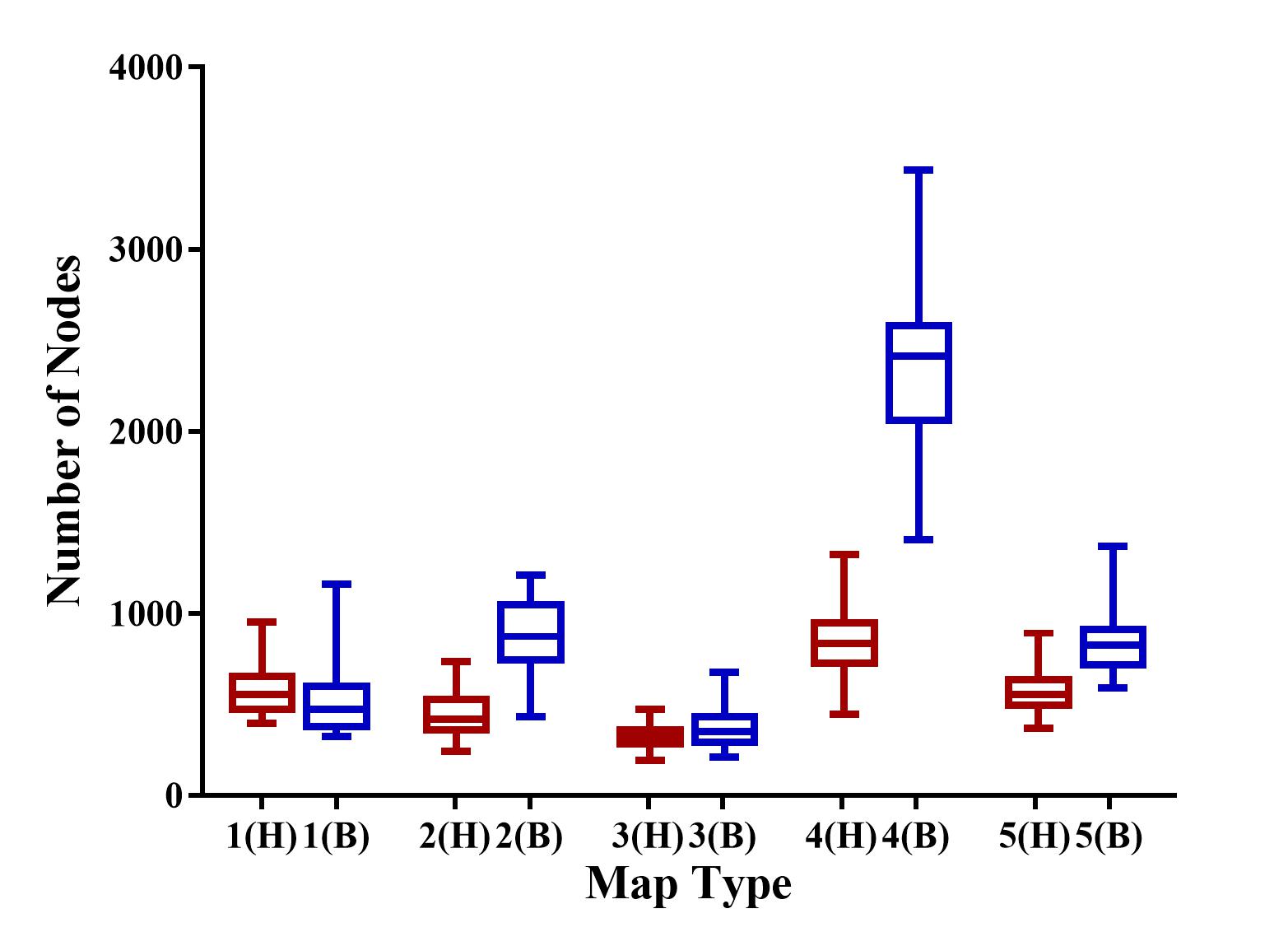}}
	\subfigure[Optimal path cost.]{
		\includegraphics[width=58mm]{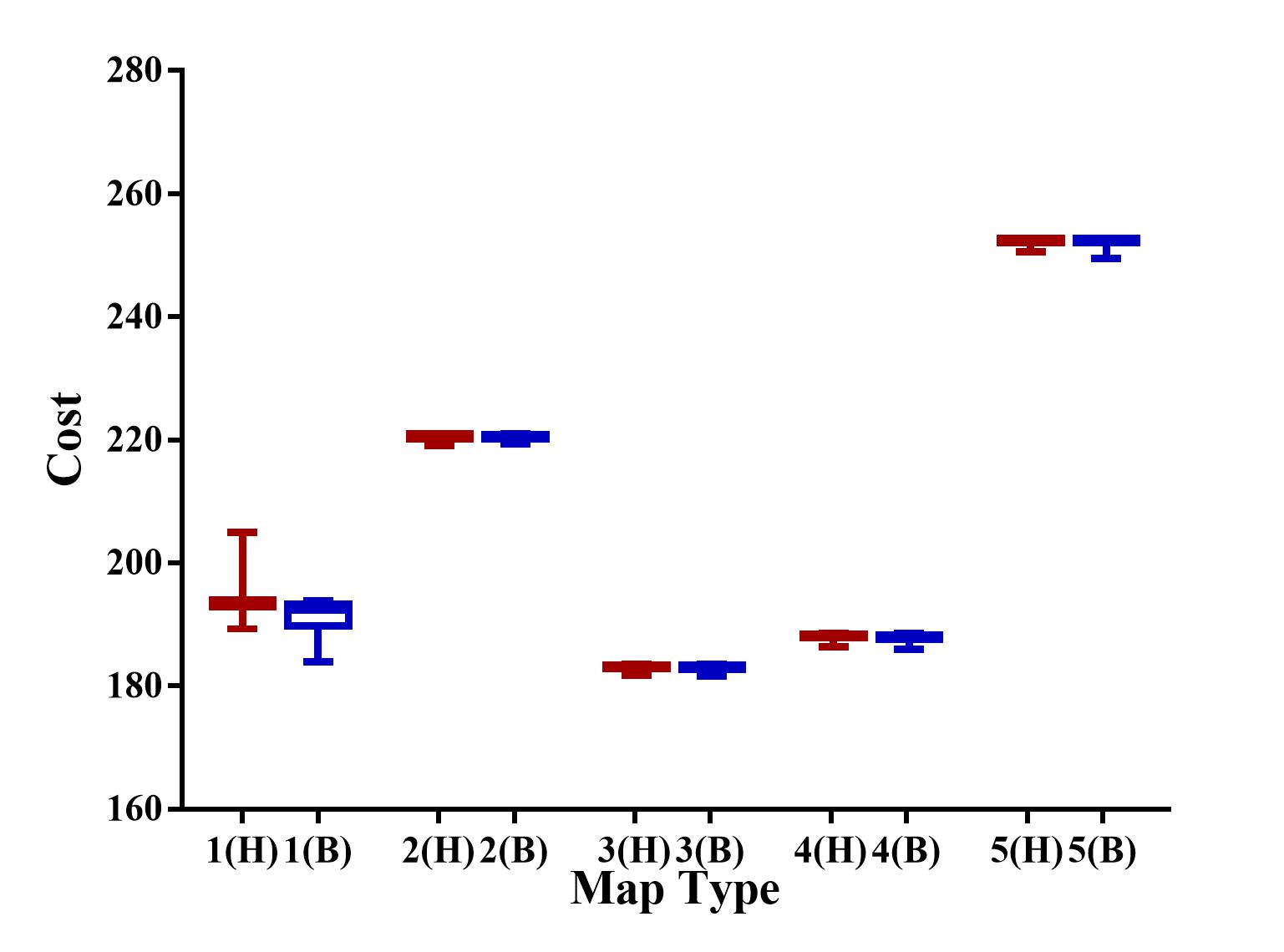}}
	\subfigure[Optimal time cost.]{
		\includegraphics[width=58mm]{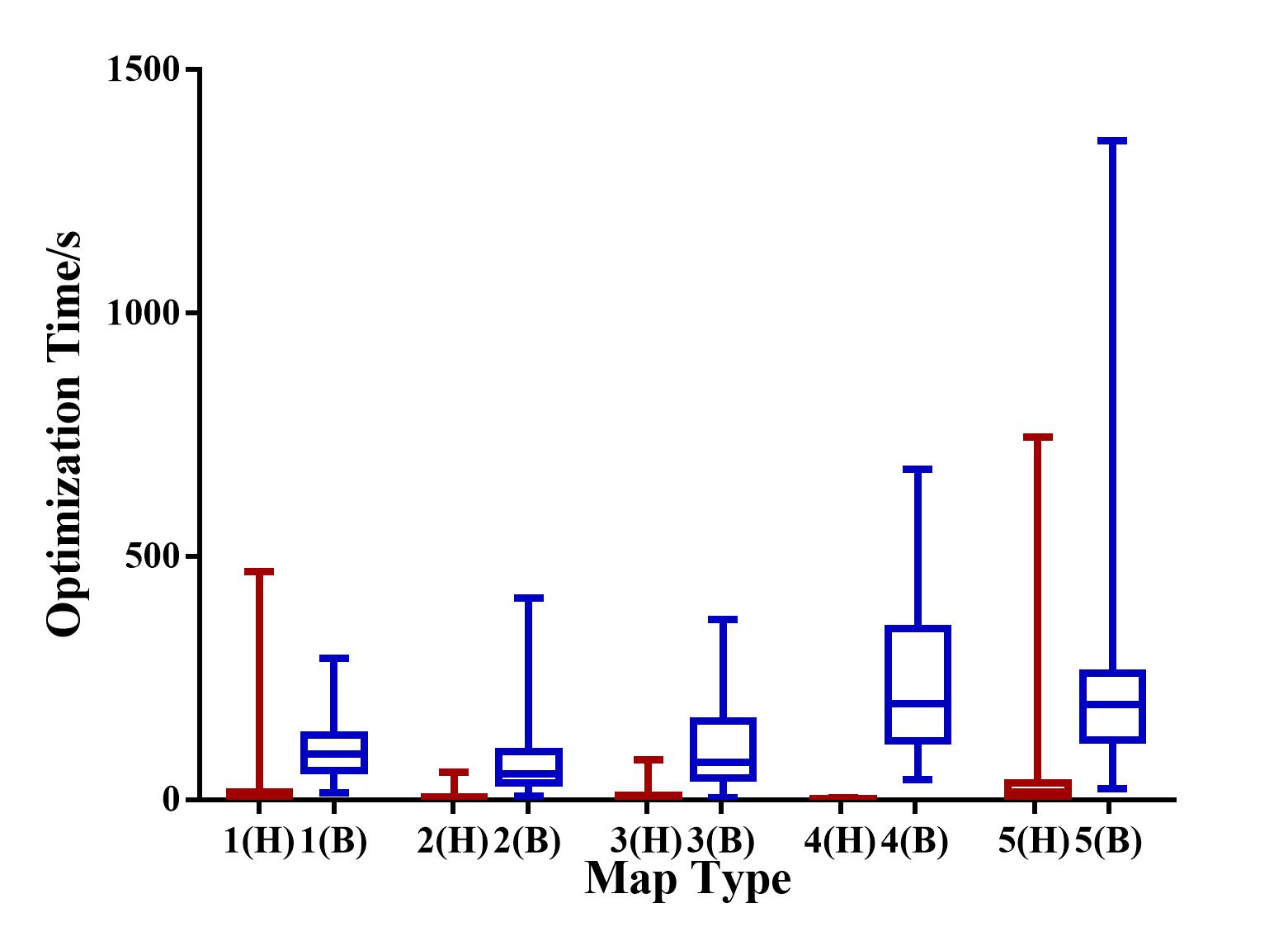}}
	\subfigure[Optimal nodes.]{
		\includegraphics[width=58mm]{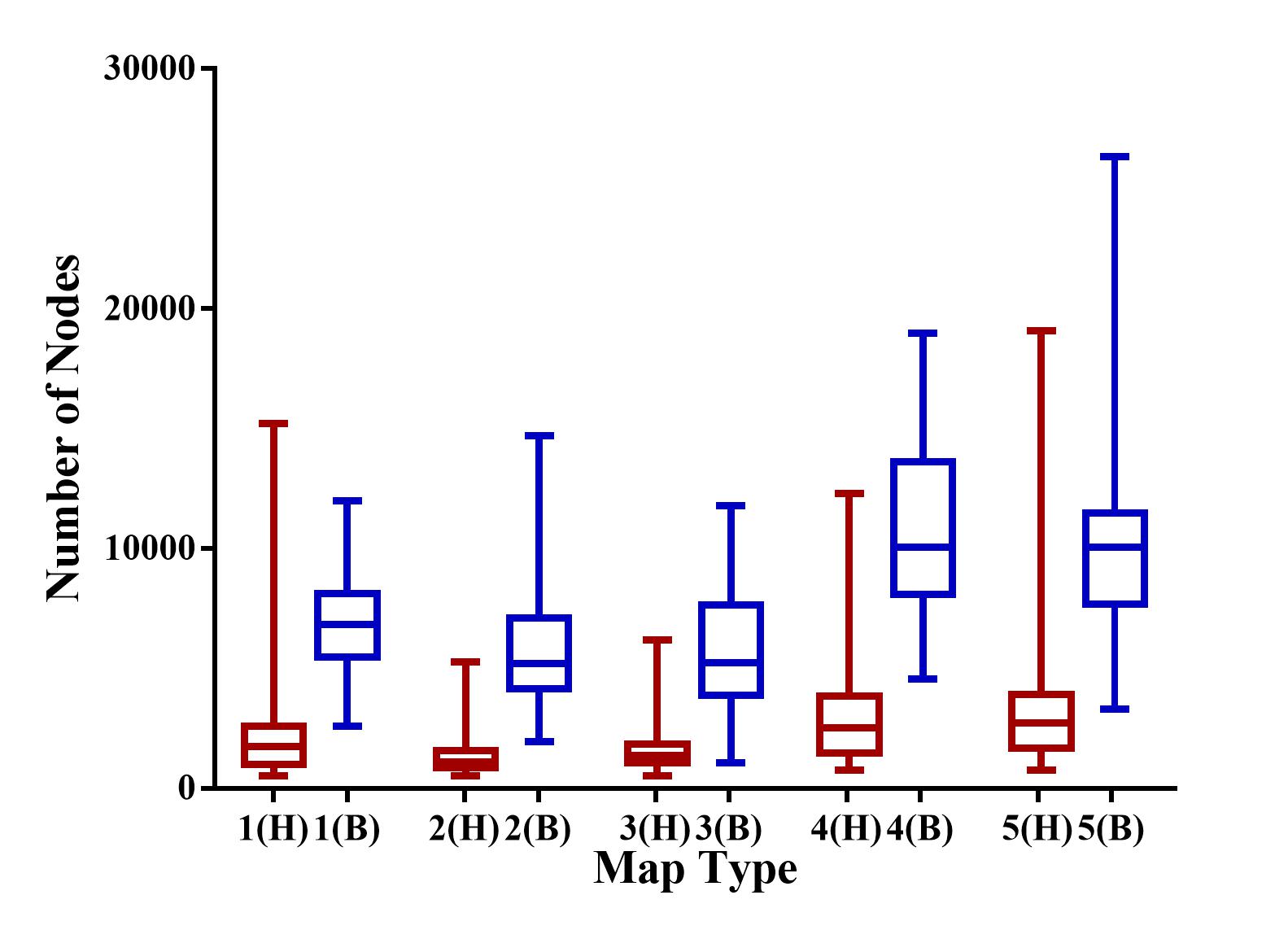}}
	\caption{Comparison of path cost, planning time, and the number of nodes between GAN-based heuristic RRT* and basic RRT* during initial stage and optimization stage. The red boxes refer to the GAN-based heuristic RRT* and the blue box indicates basic RRT*.}
	\label{fig_statistics}
\end{figure*}

Fig. \ref{fig_statistics}(a)(b)(c) displays the box-plot of the comparison on initial path cost, planing time, and the number of nodes. 
The midpoint of the boxes refers to average values and the height of boxes refers to variances.
According to the results, the GAN-based heuristic RRT* has better performance on the three indexes.
Additionally, the effectiveness is more obvious on difficult tasks (such as Map4). 
Furthermore, the heuristic non-uniform sampling distribution reduces the variance of different iterations, which means that it increases the robustness of RRT* to find a feasible path in a short time.

\textit{b. Optimization Stage}

The optimization stage refers to the period from the beginning to the time when the optimal path is found.
Fig. \ref{fig_statistics}(d)(e)(f) compares the statistical results of the three metrics in the optimization process.
As shown in Fig. \ref{fig_statistics}(d), the cost of optimal paths remains same between the two methods, which implies that placing the most of the sampling nodes of in our non-uniform sampling distribution will not affect of the cost of the optimal path.
Fig. \ref{fig_statistics}(e) and Fig. \ref{fig_statistics}(f) illustrate the significant decrease in time consumption and number of nodes: our GAN-based RRT* achieves $4$ to $12$ times faster and the number of nodes is reduced about $30\%$.

The advantage of our GAN-based heuristic RRT* is more obvious during the optimization stage, for the restriction of sampling distributions guides RRT* to avoid sampling in low possibility regions, resulting in less nodes to search during the optimization process of RRT*. 

\begin{figure}[h]
	\centering
	\includegraphics[width=88mm]{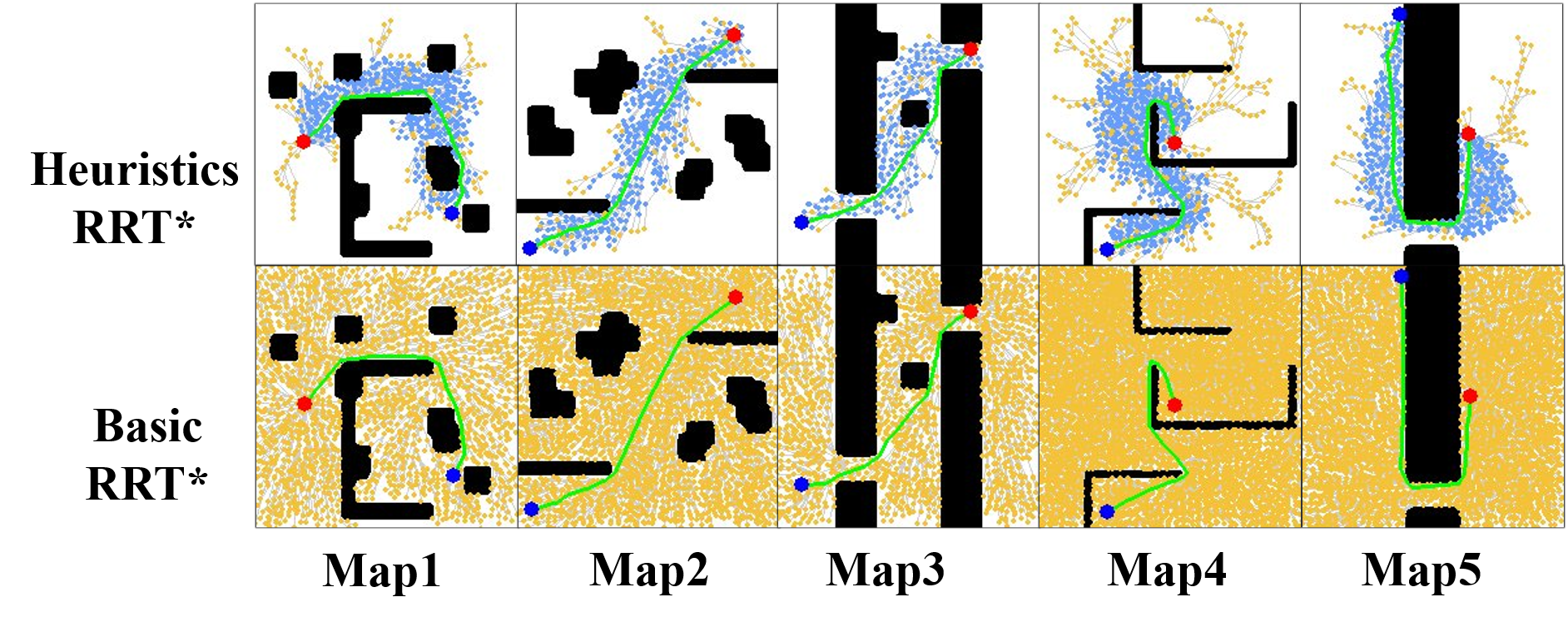}
	\caption{Comparison of RRT* expansion. The figures are the final moments of RRT*. The yellow nodes are uniformly sampled and the blue nodes are from the non-uniform sampling distributions. Starts, goals and optimal paths are marked as red, blue, green respectively.}
	\label{fig_optimal_expansion}
\end{figure}

\begin{figure}[h]
	\centering
	\subfigure[Map2 path cost.]{
		\includegraphics[width=42mm]{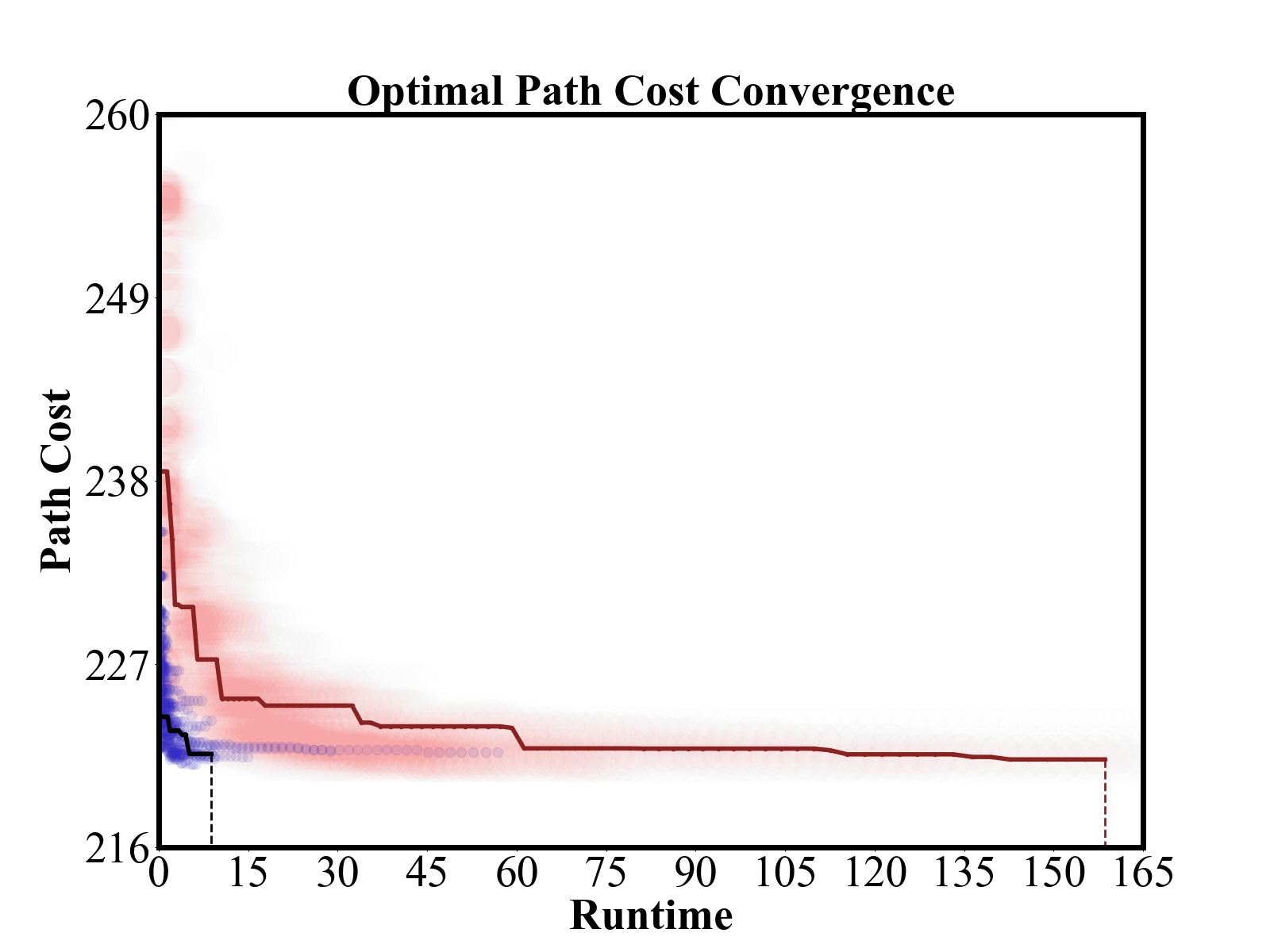}}
	\subfigure[Map2 node quantity.]{
		\includegraphics[width=42mm]{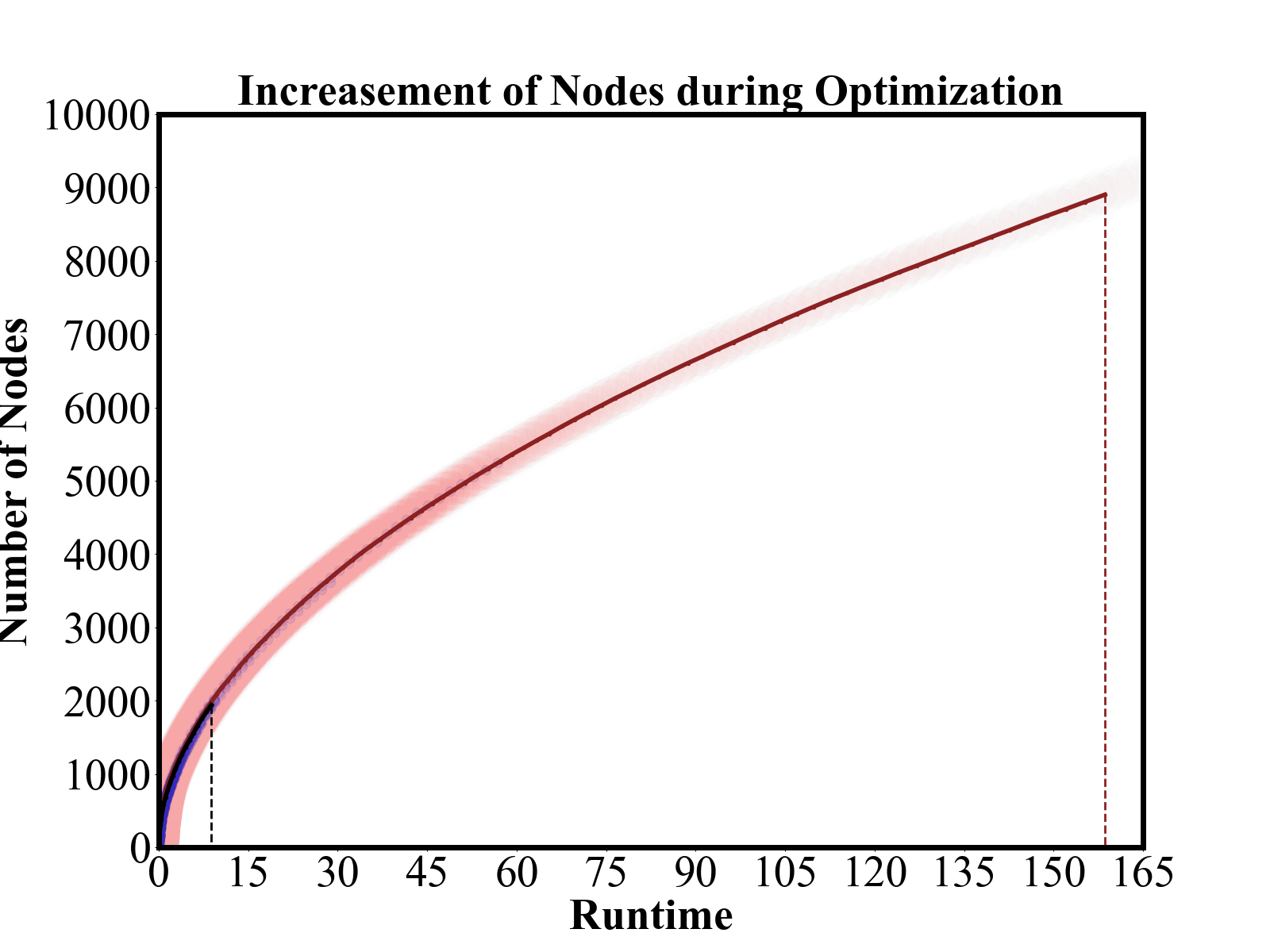}}
	\subfigure[Map4 path cost.]{
		\includegraphics[width=42mm]{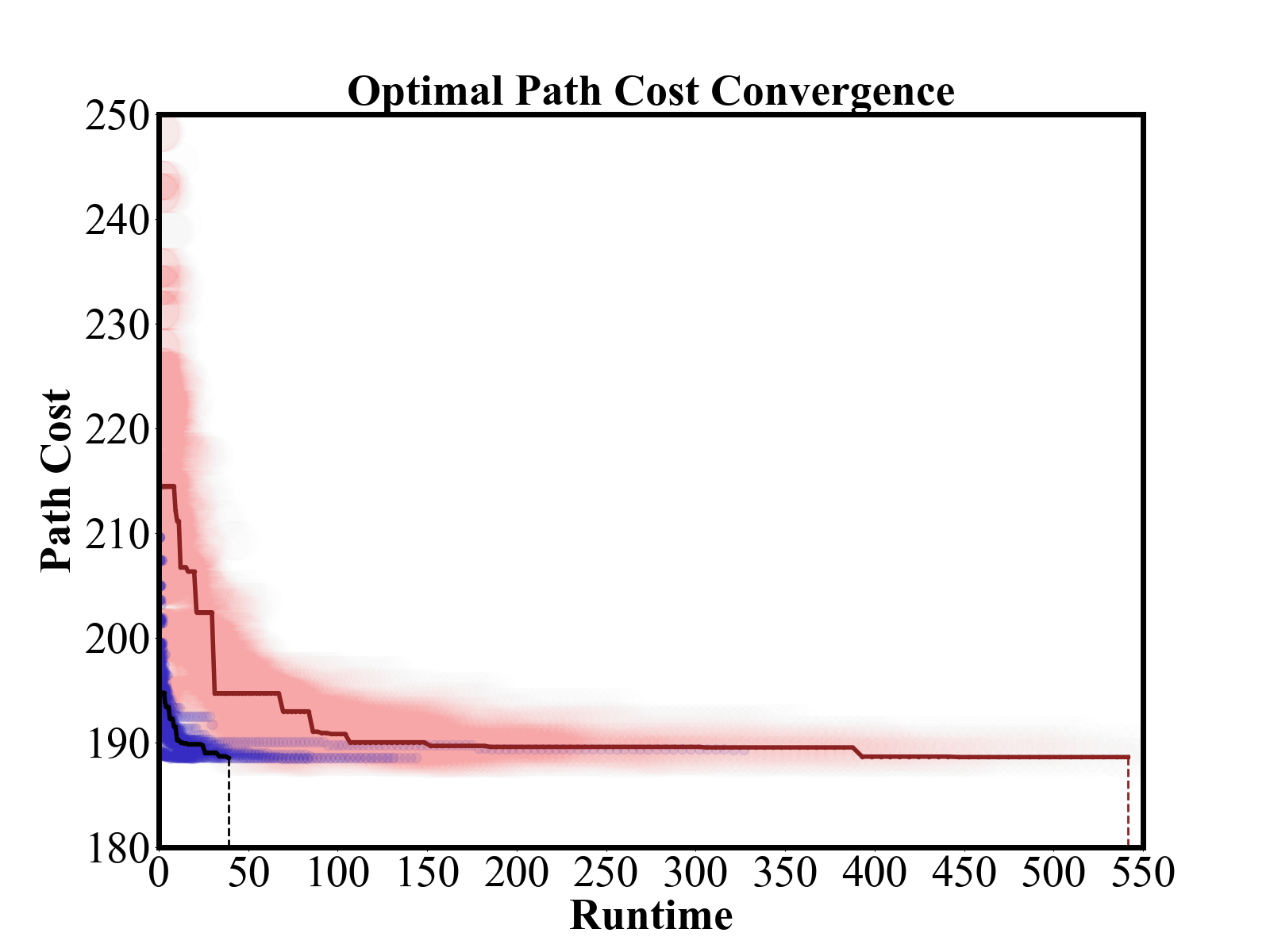}}
	\subfigure[Initial nodes quantity.]{
		\includegraphics[width=42mm]{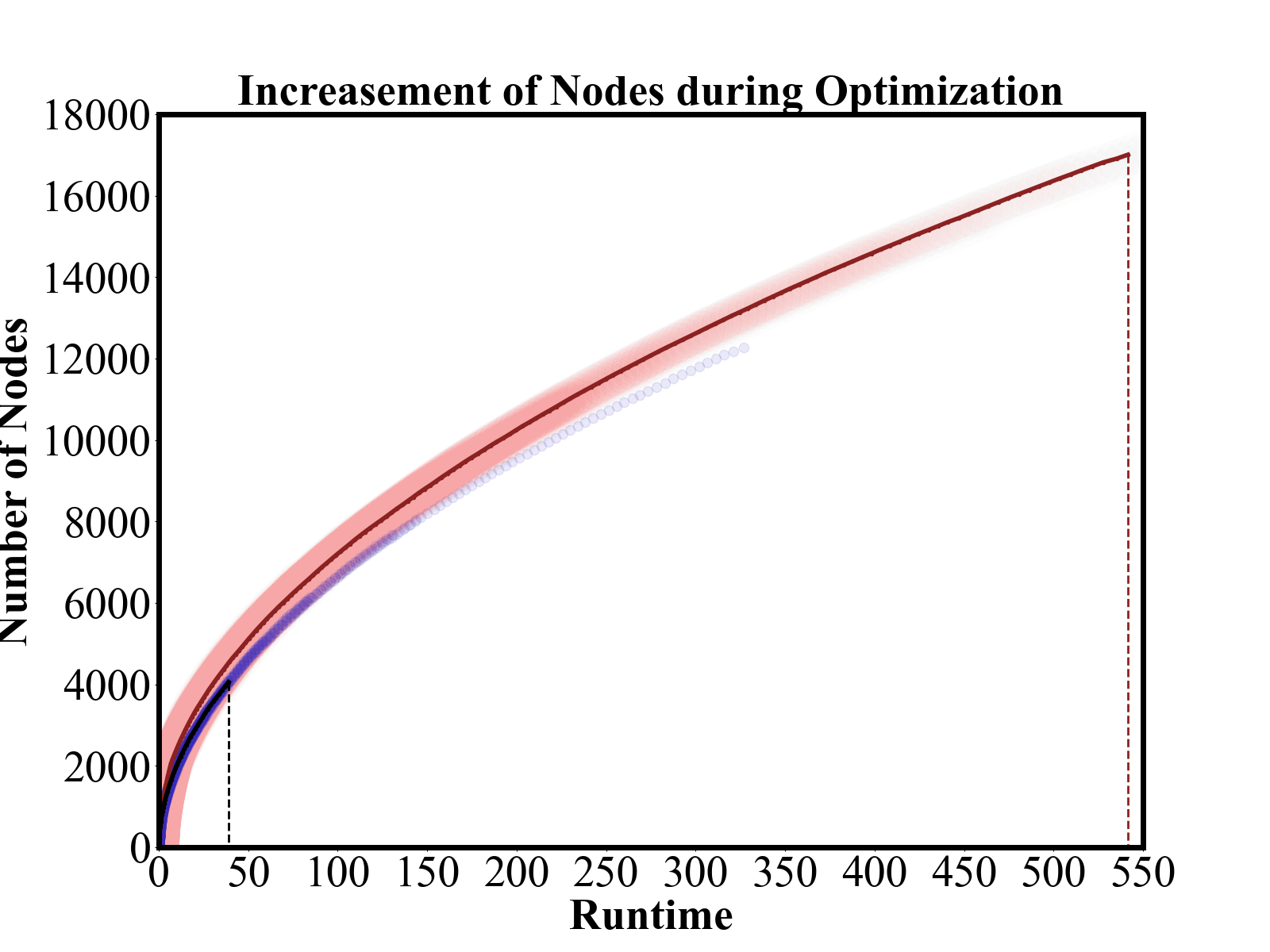}}
	\caption{Convergence process of path optimization. The blue translucent scatter refers to the multiple experiments of heuristic RRT* and the red scatters refer to basic RRT*'s. The solid line of each color is a demo from the searching process. The dotted line connecting the solid line and the x-axis marks the end moment of the searching.}
	\label{fig_optimal_examples}
\end{figure}

Fig.\ref{fig_optimal_expansion} and Fig. \ref{fig_optimal_examples} demonstrate the above-mentioned benefits more intuitively.
Fig.\ref{fig_optimal_expansion} displays the final expansion of RRT* trees when optimal paths are found.
The nodes of basic RRT* floods the whole state space.
In comparison, the number of the nodes of GAN-based heuristic RRT* are much less than the basic RRT* and distribute in the regions of high possibility.  
Fig. \ref{fig_optimal_examples} shows the convergence process of the two methods.
Fig. \ref{fig_optimal_examples}(a)(c) exhibit the convergence of path cost over time and Fig. \ref{fig_optimal_examples}(b)(d) display the increment of nodes.
The comparison demonstrates the remarkable improvement in the reduction of planning time and node quantity of our GAN-based RRT*.  

\subsection{Discussions}
Herein, we discuss some promising improvement in our current model.

As for GAN model, its architecture can be further improved in a few aspects to increase the success rate of prediction. 
For example, the current model performs imperfectly on generating distributions that cross long distances.
This is mainly because the long-distance paths account for a small ratio in the whole randomly-generated dataset, resulting in difficulties for the model to learn distant connections.
In addition, there is enough room for improvement, such as simplifying structure for lightweight network, repeating cycles for more accurate predictions, and so on. 

In terms of the implementation of sampling-based planning algorithms, heuristic rate $\mu$ may have better value to balance randomness and guidance. Furthermore, more intelligent sampling strategies can be applied to achieve heuristic non-uniform distributions to avoid redundant nodes selection when moving towards the goal.

\section{CONCLUSIONS AND FUTURE WORK}
\label{conclusion}
In this paper, we present an image-based heuristic methodology to guide non-uniform sampling-based path planning algorithms.
In particular, we design a generative adversarial network to predict promising regions and apply it to generate heuristic non-uniform sampling distributions for biased RRT* algorithm.
We evaluate the GAN-based heuristic model from the image aspect and path planning aspect.
As an image generation problem, the model shows a high success rate and strong adaptability.
As a path planning problem, the test results of our method shows significant improvement on the quality of initial paths and greatly accelerates the convergence speed to the optimum.  
Our heuristic method is more human-intuitive, which can avoid complicated preprocessing on the state space thus not confining to a specific environment.

Many researches can be continued based on the this work.
One extension is to add kinodynamic conditions (such as step size, rotation angle, etc.) as input to improve regional accuracy and enable the network to adjust promising regions according to these constraints.  
Another promising extension is to apply the model to dynamic environments. 
Specifically, the small black blocks in our maps can be considered as moving pedestrians and vehicles.
Through computing our model on each frame, it is feasible to guide sampling-based planning in dynamic environments.
The third promising work is to expand the methodology into high-dimensional configuration space to guide sampling on more complicated conditions.  

\section*{ACKNOWLEDGEMENT}
This work is partially supported by Hong Kong ITC ITSP Tier2 grant \#ITS/105/18FP, Hong Kong RGC GRF grant \# 14200618, and Hong Kong ITC MRP grant \#MRP/011/18.
We also appreciate the great help from the members of Robotics, Perception and Artificial Intelligence Lab in The Chinese University of Hong Kong.

\ifCLASSOPTIONcaptionsoff
  \newpage
\fi



%
\bibliographystyle{IEEEtran}
\bibliography{jk}

%

\begin{IEEEbiography}
	[{\includegraphics[width=1in,height=1.25in,clip,keepaspectratio]{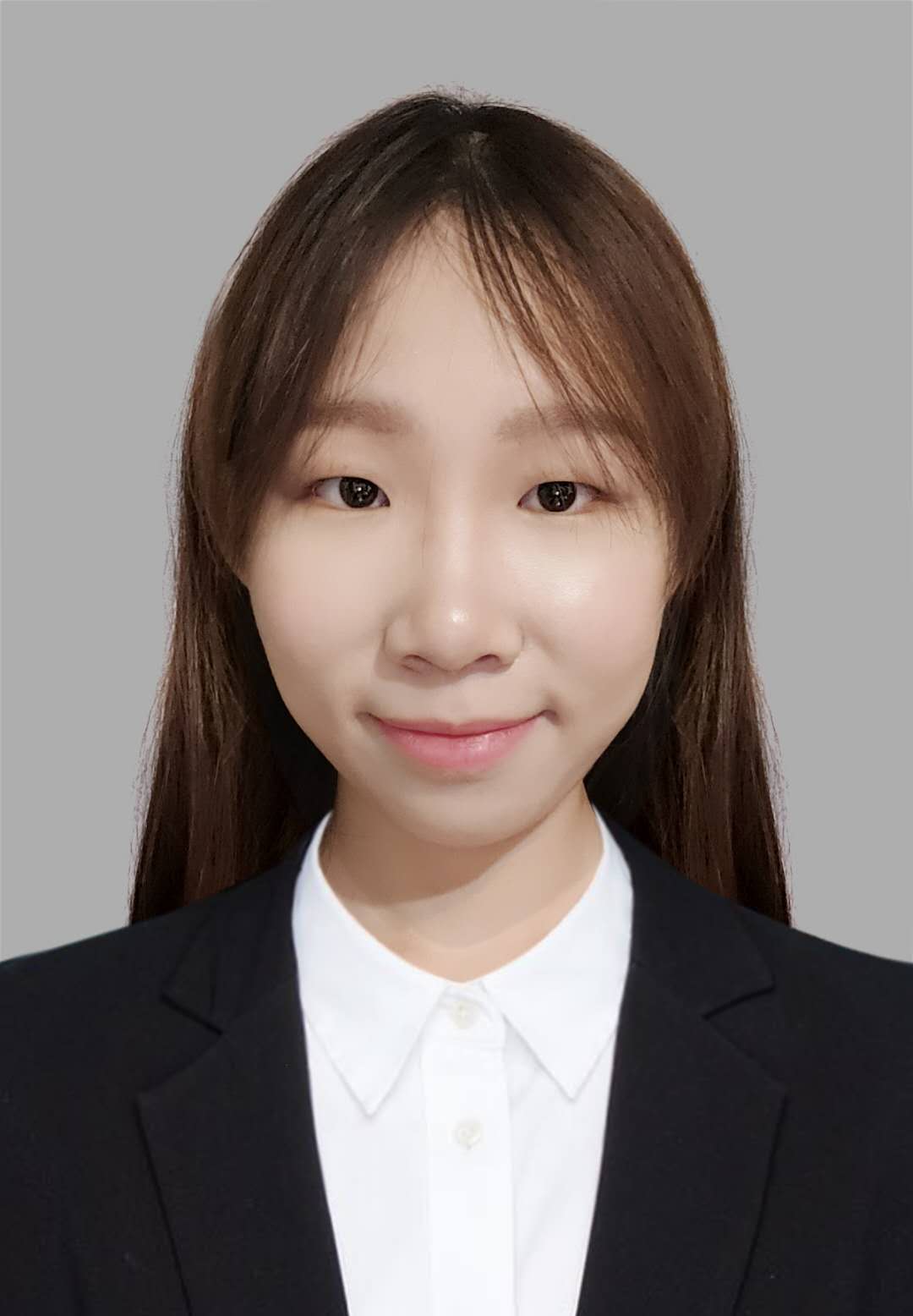}}]{Tianyi Zhang}
	received the B.E. degree in Automation from Harbin Institute of Technology, Harbin, China, in 2020.
	She is currently a research assistant with the Department of Electronic and Electrical Engineering of the Southern University of Science and Technology in Shenzhen, China.
	Her current research interests include machine learning in robotics and computer vision. 
\end{IEEEbiography}

\begin{IEEEbiography}
	[{\includegraphics[width=1in,height=1.25in,clip,keepaspectratio]{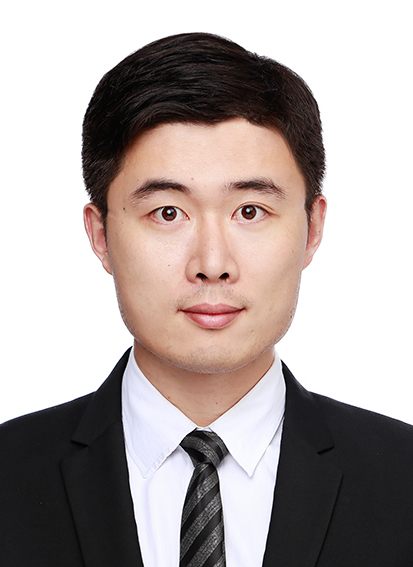}}]{Jiankun Wang}
	received the B.E. degree in Automation from Shandong University, Jinan, China, in 2015, and the Ph.D. degree in Department of
	Electronic Engineering, The Chinese University of Hong Kong, Hong Kong, in 2019. He is currently a Research Assistant Professor with the Department of Electronic and Electrical Engineering of the Southern University of Science and Technology, Shenzhen, China.
	
	During his Ph.D. degree, he spent six months at Stanford University, CA, USA, as a Visiting Student Scholar supervised by Prof. Oussama Khatib. His current research interests include motion planning and control, human robot interaction, and machine learning in robotics.
\end{IEEEbiography}

\begin{IEEEbiography}
	[{\includegraphics[width=1in,height=1.25in,clip,keepaspectratio]{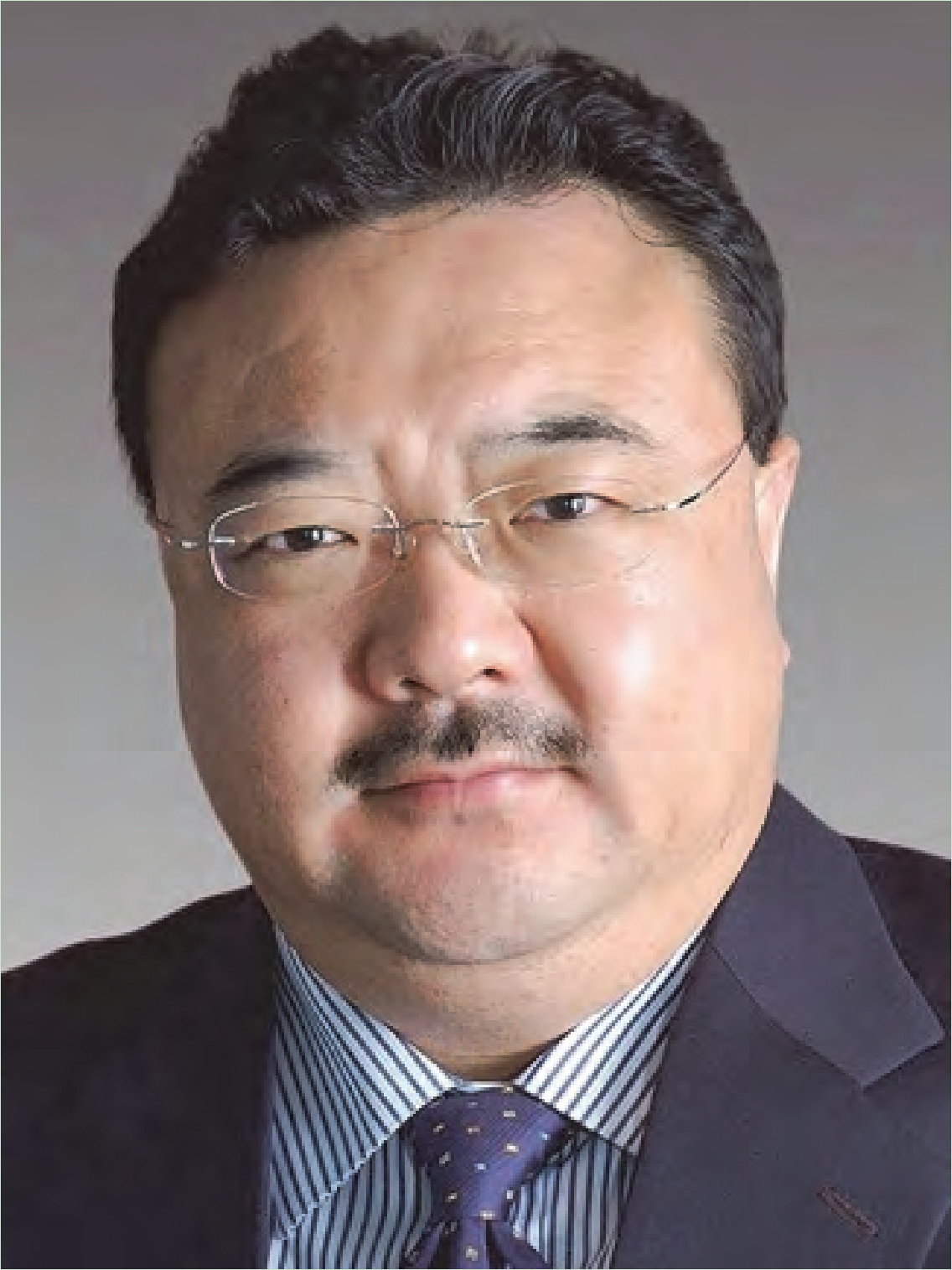}}]{Max Q.-H. Meng}
	received the Ph.D. degree in electrical and computer engineering from the University of Victoria, Victoria, Canada, in 1992.
	
	He is a chair professor with the Department of Electronic and Electrical Engineering of the Southern University of Science and Technology in Shenzhen, China, on leave from the Department of Electronic Engineering, The Chinese University of Hong Kong, Hong Kong, and also with the Shenzhen Research Institute of the Chinese University of Hong Kong in Shenzhen, China.
	He holds honorary positions as a Distinguished Professor with State Key Laboratory of Robotics and Systems, Harbin Institute of Technology, Harbin,
	China; a distinguished Provincial Professor with Henan University of Science and Technology, Luoyang, China; and the Honorary Dean of the School of Control Science and Engineering, Shandong University, Jinan, China. 
	His research interests include robotics, perception and sensing, human–robot interaction, active medical devices, biosensors and sensor networks, and adaptive and intelligent systems. He has published more than 500 journal and conference papers and served on many editorial boards.
	
	Dr. Meng is serving as an Elected Member of the Administrative Committee of the IEEE Robotics and Automation Society. He received the IEEE Third Millennium Medal Award.
\end{IEEEbiography}








\end{document}